\documentclass[runningheads]{llncs}

\usepackage{eccv}

\usepackage{xcolor}

\usepackage{eccvabbrv}

\usepackage{graphicx}
\usepackage{booktabs}
\usepackage{multicol}
\usepackage{multirow}
\usepackage{array}
\usepackage[normalem]{ulem}
\usepackage{subcaption}
\newcolumntype{?}{!{\vrule width 1pt}}
\usepackage[accsupp]{axessibility}  %
\usepackage{wrapfig}

\usepackage{hyperref}

\usepackage{orcidlink}

\definecolor{RDcolor}{rgb}{0.5, 0.1, 0.8}

\begin{document}

\title{Generative Relightable Avatars}

\author{Kunwar Maheep Singh \and
Christian Theobalt \and
Rishabh Dabral}

\authorrunning{K.M.~Singh et al.}

\institute{Max Planck Institute for Informatics,\\Saarland Informatics Campus\\
\email{\{ksingh,theobalt,rdabral\}@mpi-inf.mpg.de}}

\maketitle

\vspace{-12pt}
\begin{abstract}
We present Generative Relightable Avatars (GRA), a person-specific method for photorealistic free-view rendering and environment-map relighting of full-body humans. 
We postulate that modeling fine-grained appearance details is inherently a one-to-many problem that can benefit from a generative formulation.
In contrast to fully regressive relightable avatar methods, GRA follows a hybrid approach that combines controllable, physics-grounded relighting with probabilistic refinement.
Starting from a tracked animated mesh, we optimize material parameters in UV-space and render a coarse relit appearance under a target HDR environment map.
Next, we refine the textures with a feed-forward model  to capture pose-dependent texture dynamics and illumination effects beyond simplified reflectance assumptions.
Finally, a fine-tuned video-to-video diffusion model transforms the physically grounded renderings into temporally coherent, high-detail videos while preserving 3D control, with an error-recycling strategy for generating long videos. 
Experimental evaluations demonstrate our method's improved perceptual quality over prior relightable avatar baselines.
We urge the readers to watch the supplementary video. See the \href{https://vcai.mpi-inf.mpg.de/projects/GRA/}{project page} for more details.

\end{abstract}
\vspace{-30pt}
\section{Introduction} \label{sec:introduction}
\vspace{-8pt}
\par 
Developing photorealistic digital human avatars that can be realistically relit under arbitrary illumination remains a long-standing goal in computer vision and graphics, central to applications in telepresence, virtual production, and AR/VR. The challenge is particularly acute for full-body, clothed humans under dynamic motions, where minute appearance cues like wrinkles and shading critically influence realism.

\par
Prior works on relightable 3D avatars largely fall into two categories.
One line of work comprises deterministic pipelines that reconstruct a controllable 3D human representation, estimate explicit reflectance factors, and render novel views under novel illumination via physically based rendering or a feed-forward model~\cite{chen2022relighting4d,chen2024meshavatar,wang2024intrinsicavatar,xu2024relightable,relightneuralactor2024eccv,wang2025relightable,jiang2025dnf}.
While such methods provide strong 3D controllability, their inherent determinism and limited reflectance modeling restrict their ability to capture the one-to-many relationship between skeletal motion and fine appearance details, leading to over-smooth renderings.
In contrast, a second line of methods adopts generative relighting in the image or video domain, producing convincing lighting changes~\cite{he2024diffrelight,mei2025lux} but typically lacking physical consistency and explicit 3D control, limiting their applicability for free-viewpoint rendering.
\begin{figure}[t!]
\centering

\includegraphics[trim={21cm 2.7cm 5.4cm 16cm},clip,width=\textwidth]{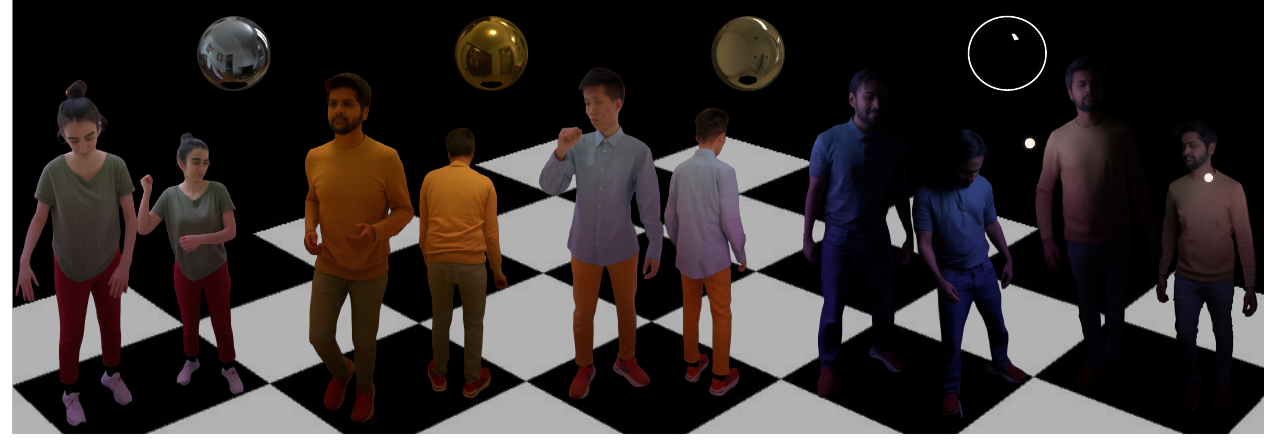} 

\caption{We present \emph{Generative Relightable Avatars}, a novel method that enables photorealistic relighting of full-body humans in motion. From pose, viewpoint, and illumination as input, our method generates temporally consistent, physically grounded renderings under novel lighting, which also qualitatively generalizes to out-of-distribution one-light-at-a-time (OLAT) setups and near-field lighting effects. This enables the realistic placement of dynamic humans into arbitrary virtual environments.}
\vspace{-20pt}
\label{fig:teaser}
\end{figure}
\par
To address these challenges, we introduce Generative Relightable Avatars (GRA), a relightable framework for generating pose-driven photorealistic renderings under arbitrary viewpoints and illumination. Rather than asking a single generative model to jointly infer geometry, motion, viewpoint, and lighting, we adopt a coarse-to-fine decomposition: (i) an explicit 3D avatar provides controllable, physically grounded intermediate renderings under the target environment map, and (ii) a conditional video generative model probabilistically synthesizes the fine-scale details that are inherently stochastic and ambiguous under imperfect tracking, as shown in Fig.~\ref{fig:teaser}.

\par 
Concretely, we first train a person-specific parametric avatar that maps tracked 3D motion and camera parameters to a temporally consistent animated mesh. We optimize person-specific material parameters in UV space and render the mesh under a target environment map using a microfacet reflectance model~\cite{cook1982microfacet}. A lightweight neural model, RelightNet, then captures illumination effects beyond the simplified reflectance assumptions and models pose-dependent texture dynamics in UV space. By construction, this stage supports novel viewpoint synthesis, yielding physically grounded renderings that capture the overall lighting response but remain visually coarse.
\par
In the second stage, we finetune a video generative model to resolve the one-to-many ambiguity in high-frequency appearance by formulating it as a video-to-video task over the coarse physically-based renderings. 
We introduce an Atemporal Cross-Attention block to inject illumination conditioning 
by cross-attending per-frame environment maps with the corresponding video frames. 
This encodes per-frame lighting correspondence as an inductive bias, motivated
by our training data where every frame is captured under a distinct
illumination. 
To support long dynamic motions beyond the limited generation horizon of current video diffusion models, we integrate an error-recycling strategy~\cite{li2025stable} that maintains temporal consistency over substantially longer sequences.

In summary, our contributions are threefold.
\begin{itemize}
    \item We propose a system for  relightable human avatars that combines physical plausibility with photorealism through a hybrid explicit and generative pipeline, enabling the sampling of diverse wrinkle patterns.%
    \item A video relighting generative model that introduces novel lighting controls to a video-to-video generative model, while ensuring long-form video generation %
    \item We achieve state-of-the-art results with qualitative out-of-distribution performance on unseen environment illumination, including OLAT sequences.
\end{itemize}
\vspace{-18pt}

\section{Related Work}
\vspace{-6pt}
\subsection{3D Full-body Relightable Avatars.}
Recovering a relightable full-body human avatar from images has been widely studied, with approaches differing primarily in their assumptions about illumination calibration, sensor setup, and appearance representation.
Most methods formulate the problem as inverse rendering, where geometry, material, and illumination parameters are optimized through a differentiable rendering pipeline to match observed images.

\smallskip
\par \noindent \textbf{Inverse Rendering under Uncalibrated Illumination.}
A large body of work targets relightable avatar reconstruction from casually captured images under unknown lighting~\cite{chen2022relighting4d, chen2024meshavatar, wang2024intrinsicavatar, xu2024relightable, xie2024ren, zhao2024surfel, iqbal2023rana, jiang2025dnf}, where single-view or multi-view observations are used to jointly estimate environment illumination along with dynamic geometry and material properties. However, recovering illumination jointly with material and geometry introduces severe ambiguities, often necessitating strong priors or simplified reflectance models. To make the problem tractable, most approaches adopt analytic BRDFs such as microfacet~\cite{chen2022relighting4d, relightneuralactor2024eccv}, Disney~\cite{wang2024intrinsicavatar}, or Lambertian models~\cite{iqbal2023rana, xiao2024neca}, and represent appearance using implicit neural fields defined in a canonical pose and deformed into observation space via skinning~\cite{chen2022relighting4d}, enabling differentiable rendering but complicating accurate visibility reasoning. Subsequent works improve shading fidelity by combining explicit meshes with implicit material fields~\cite{chen2024meshavatar}, allowing more reliable surface normals and view-dependent effects. Significant effort has also been devoted to accelerating visibility and light transport computation; for example, Xu et al.~\cite{xu2024relightable} introduce hierarchical distance queries to efficiently compute ray–surface intersections in deforming implicit fields, while Zhao et al.~\cite{zhao2024surfel} use surfel-based representations to speed up visibility estimation. Despite these advances, uncalibrated inverse rendering methods often suffer from residual lighting–material ambiguity, and render over-smoothed appearance.

\smallskip
\par \noindent \textbf{Relighting with Calibrated Illumination.}
To alleviate illumination ambiguity, several methods assume partially or fully calibrated lighting during capture.
Relightable Neural Actor~\cite{relightneuralactor2024eccv} leverages a small set of known illumination conditions, simplifying optimization and improving shading consistency.
More recently, Relightable Full-body Gaussian Codec Avatars~\cite{wang2025relightable} exploit a lightstage setup with hundreds of illumination conditions, obtained via randomized groupings of OLAT measurements, to densely sample the illumination basis space and tightly constrain material estimation.
While calibrated lighting significantly improves physical correctness, these methods still rely on deterministic appearance models and remain sensitive to tracking errors and limited pose coverage.

\vspace{-12pt}
\subsection{Generative Relighting Methods.}
\vspace{-6pt}
Generative priors have been transformative in estimating the geometry~\cite{trellis, hunyuan}, and material~\cite{trellis2, rgbx} properties of objects from input images.
Image-space diffusion models have been successfully adapted to perform scene relighting ~\cite{jin2024neural,zhang2024iclight,erel2025practilight}, portrait relighting~\cite{pandey2021total,kim2024switchlight,mei2024holo,chaturvedi2025synthlight} and associated tasks such as shadow editing and harmonization \cite{yoon2024generative,ren2024relightful}.
More recent works like Comprehensive Relighting address the broader human setting with temporal input and background harmonization \cite{wang2025comprehensive}.
Closely related to our setting are the works that perform video relighting via video diffusion models or diffusion-based neural rendering \cite{mei2025lux,wang2025relumix,liang2025diffusionrenderer}.
While these approaches produce highly photorealistic lighting effects, they typically operate in the image plane and, therefore, offer limited explicit 3D control.
In contrast, our work couples an explicit, pose- and view-controllable full-body relightable avatar with a conditional video generative refinement model.

\vspace{-8pt}

\section{Method} \label{sec:method}
\vspace{-8pt}
Our goal is to perform physics-guided, photorealistic relighting of a human character performing motion $\boldsymbol{\theta}$, rendered from arbitrary viewpoint $\boldsymbol{\kappa}$ under arbitrary environment lighting $\mathbf{E}$. 
As motion is a sparse input signal, 
rendering fine-grained appearance details such as wrinkles is inherently a one-to-many task: the same skeletal motion can explain a variety of wrinkle patterns and appearance effects.
This motivates us to seek a generative solution to the task, where the character's appearance, $\mathbf{z}$, can be modeled as a conditional distribution $p(\mathbf{z} | \boldsymbol{\theta}, \boldsymbol{\kappa}, \mathbf{E})$ over the skeletal motion, camera viewpoint and the environment illumination.
\par
However, jointly learning the conditional probability distribution above is challenging as the model needs to precisely disentangle the effects of illumination, viewpoint and motion.
Therefore, we follow a divide-and-conquer approach and reformulate the task as $p(\mathbf{z} | \boldsymbol{\theta}, \boldsymbol{\kappa}, \mathbf{E}) = p(\mathbf{z} | \mathbf{z}^{\prime}, \mathbf{E})$, where $\mathbf{z}^{\prime} = q(\boldsymbol{\theta}, \boldsymbol{\kappa}, \mathbf{E})$ is a deterministic function representing the intermediate, coarse renderings of a mesh avatar illuminated with $\mathbf{E}$ and rendered from the target viewpoint $\boldsymbol{\kappa}$.
This simplifies the generative task to that of modeling the fine-grained details like wrinkles, and generating realistic illumination-dependent effects like shadows and specularity from coarse guidance.
In practice, we model $p(\mathbf{z} | \mathbf{z}^{\prime}, \mathbf{E})$ by adapting a video-to-video generative model to accept illumination conditioning. %
The intermediate representation $\mathbf{z}^{\prime} = q(\boldsymbol{\theta}, \boldsymbol{\kappa}, \mathbf{E})$ is modeled by taking a person-specific 3D avatar~\cite{habermann2021real} and making it relightable by optimizing material properties under approximate reflectance assumptions, followed by learning pose and illumination-dependent effects with a feed-forward neural network.
\par
Fig.~\ref{fig:main} outlines our overall schema.
Following a coarse-to-fine strategy, we first use a mesh-based 3D avatar that predicts a coarse surface from skeletal motion. On this surface, an explicit physics-based texture produces an approximate relighting of the character under novel illumination (Sec.~\ref{sec:microfacet}). A subsequent feed-forward network, RelightNet, then approximates illumination-dependent effects beyond the coarse microfacet approximation,
yielding a coarse, but physically grounded relit appearance (Sec.~\ref{sec:relightnet}). 
Fine-scale details are modeled probabilistically using a generative model for refinement (Sec.~\ref{sec:gen_refine}), capturing the inherent ambiguity in mapping pose to high-frequency appearance variations.
\vspace{-12pt}
\subsection{Physics-based Texture} \label{sec:microfacet}
\vspace{-8pt}

\begin{figure*}[t]
\centering
    \includegraphics[width=\textwidth]{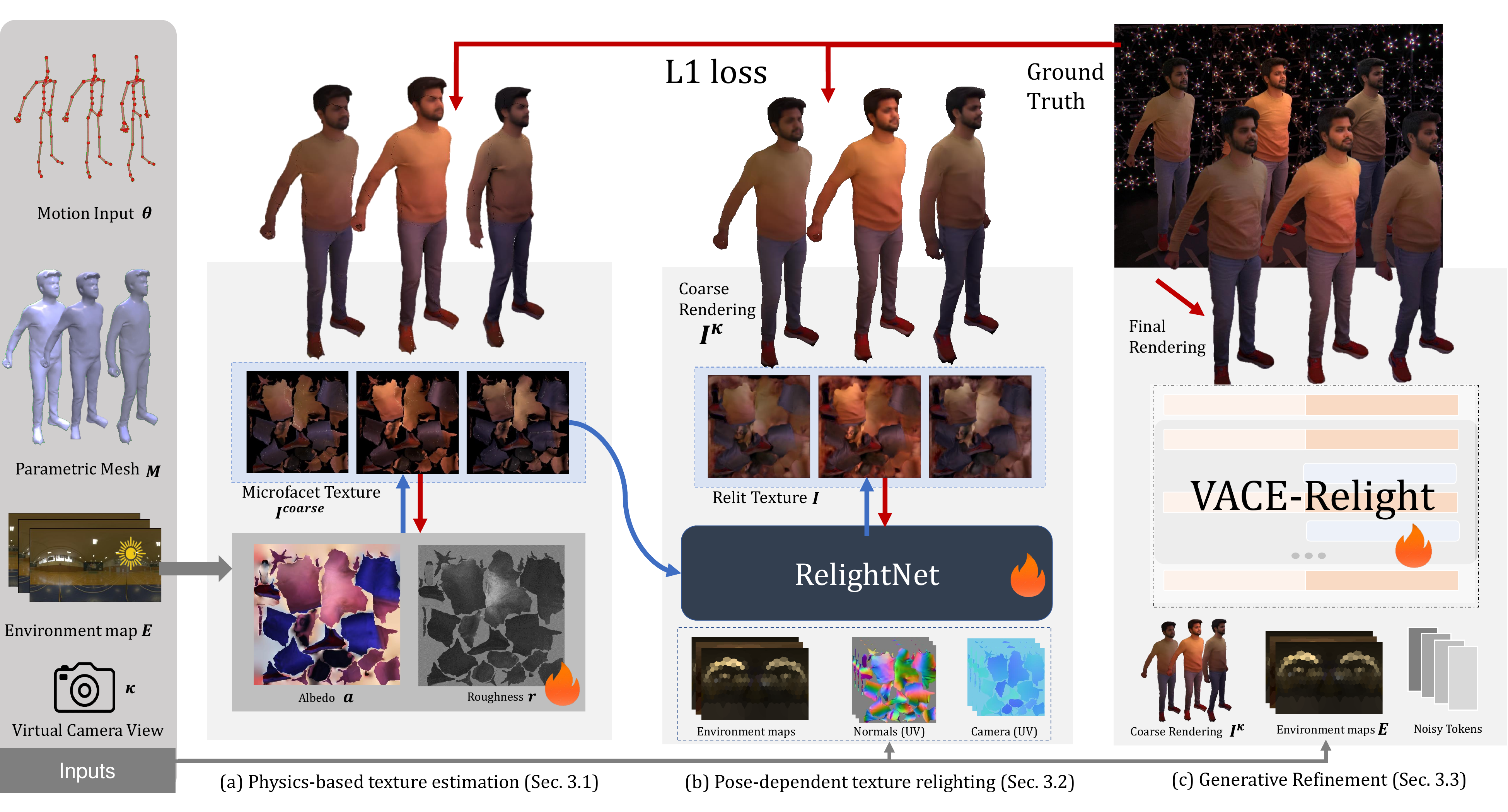}
    \vspace{-12pt}
    \caption{\textbf{Overview.} Given skeletal motion $\boldsymbol{\theta}$, viewpoint
$\boldsymbol{\kappa}$, and environment illumination $\mathbf{E}$, GRA outputs a photorealistic rendering of the character avatar in three steps:
(a) A person-specific animatable mesh predicts
pose-dependent geometry $M$, relit under a microfacet BRDF to produce a coarse
appearance (Sec.~\ref{sec:microfacet}), which is refined by (b) RelightNet to
capture illumination- and pose-dependent effects (Sec.~\ref{sec:relightnet}).
(c) Finally, a finetuned video diffusion model  synthesizes
fine-scale details conditioned on the coarse renderings and per-frame
environment maps %
(Sec.~\ref{sec:gen_refine}). 
Blue arrows indicate the inference path; red
arrows indicate gradient propagation.}
     \vspace{-12pt}
    \label{fig:main}
\end{figure*}

We begin with a person-specific parametric body model~\cite{habermann2021real} that is trained to produce a temporally consistent character mesh $M_t$, given the input skeletal motion $\boldsymbol{\theta}_t$ at time $t$.
The parametric model is also trained to produce dynamic, view-dependent textures, but is agnostic to the target illumination $\mathbf{E}$.
Therefore, our first goal is to obtain a physics-guided approximation of the character's appearance under novel lighting that can effectively guide subsequent \textit{learned} relighting.
To this end, we introduce an explicit, physically motivated texture that approximates the rendering equation for a deforming human surface.

The outgoing radiance $\boldsymbol{L}_o$ at a surface point $\boldsymbol{x}$ with outgoing direction $\boldsymbol{\omega}_o$ is given by the rendering equation~\cite{kajiya1986rendering}:
\begin{equation}
\boldsymbol{L}_o(\boldsymbol{x}, \boldsymbol{\omega_o}) = \int_{\boldsymbol{\omega}_i} f_r(\boldsymbol{x},\boldsymbol{\omega}_i,\boldsymbol{\omega}_o)
\boldsymbol{L}_i(\boldsymbol{x},\boldsymbol{\omega}_i)
\boldsymbol{V}(\boldsymbol{x,\omega}_i)
\langle \boldsymbol{\omega}_i, \boldsymbol{n}\rangle  d\boldsymbol{\omega}_i,
\label{eq:rendering_eq}
\end{equation}
where $\boldsymbol{\omega}_i$ is the incoming light direction, $\boldsymbol{n}$ is the normal at $\boldsymbol{x}$, $f_r$ is the surface BRDF, $\boldsymbol{L}_i$ is the incoming light intensity and $\boldsymbol{V}$ indicates visibility.
We model the surface reflectance using a simplified Cook--Torrance microfacet BRDF~\cite{cook1982microfacet}, parameterized by spatially varying albedo $\boldsymbol{a}$ and roughness $\boldsymbol{r}$ maps:
\begin{equation}
f_r(\boldsymbol{x}, \boldsymbol{\omega}_i, \boldsymbol{\omega}_o)
=
\mathrm{BRDF}(\boldsymbol{\omega}_i, \boldsymbol{\omega}_o; \boldsymbol{a}(\boldsymbol{x}), \boldsymbol{r}(\boldsymbol{x})).
\end{equation}

As the albedo and roughness maps are unknown, we optimize for them by evaluating
Eq.~\ref{eq:rendering_eq} in the UV space of $M_t$ under a one-bounce
assumption using the known mesh geometry, camera parameters $\boldsymbol{\kappa}$,
and environment map $\mathbf{E}$:
\begin{equation}
\boldsymbol{a}^*, \boldsymbol{r}^* = \argmin_{\boldsymbol{a}, \boldsymbol{r}}
\sum_t \left\|
\mathcal{P}\!\left(M_t, \boldsymbol{\kappa}, \mathbf{E};
\boldsymbol{a}, \boldsymbol{r}\right) - \boldsymbol{I}_t^{\mathrm{gt}}
\right\|_1
\label{eq:argmin}
\end{equation}
where $\mathcal{P}(\cdot)$ denotes the differentiable rendering of the mesh
under the microfacet BRDF, and $\boldsymbol{I}_t^{\mathrm{gt}}$ is the
ground-truth image. This optimization yields a coarse UV-space texture
$\mathbf{I}_t^{\text{coarse}}$ under arbitrary illumination
(see Fig.~\ref{fig:main}).

While approximate, $\boldsymbol{I}^{\mathrm{coarse}}_t$ encodes physically meaningful dependencies on geometry, view direction, and illumination, and serves as a strong physics-guided inductive bias for subsequent learned relighting.
\vspace{-8pt}
\subsection{RelightNet} \label{sec:relightnet}
The explicit physics-based texture $\mathbf{I}_t^{\text{coarse}}$ estimated above provides a physically grounded but coarse approximation of surface appearance as can be seen in Fig.~\ref{fig:main}.
This approximation has two key limitations: (i) due to the restricted expressiveness of the microfacet BRDF and the single-bounce assumption, it does not span the full space of relit appearances observed in real data; and (ii) the use of a static UV texture on an imperfectly tracked deforming mesh leads to pose-dependent smoothing of high-frequency appearance.

To address these limitations, we introduce \emph{RelightNet} $\mathcal{R}$, a feed-forward neural model that approximates the complete rendering equation in the UV space of the tracked mesh. RelightNet learns a mapping from illumination, geometry, and view configuration to the final relit mesh texture $\boldsymbol{I}_t$ as:
\begin{equation}
\boldsymbol{I}_t
=
\mathcal{R}
\big(
\mathbf{E}_t,
\boldsymbol{I}^{\mathrm{coarse}}_t,
\boldsymbol{N}_{t-k:t},
\boldsymbol{C}_t
\big),
\end{equation}
where $\mathbf{E}_t$ denotes the environment illumination, $\boldsymbol{I}^{\mathrm{coarse}}_t$ is the coarse relit texture from Sec.~\ref{sec:microfacet}, $\boldsymbol{N}_{t-k:t}$ is the set of surface normal maps with a  history of $k$ frames, and $\boldsymbol{C}_t$ encodes the per-texel view direction to the camera center. The relit texture $\boldsymbol{I}_t$ is projected back onto the mesh and rasterized into camera $\boldsymbol{\kappa}$ to get the relit rendering $\boldsymbol{I}_t^{\boldsymbol{\kappa}}$. 

\begin{figure}[t]
    \centering
    \includegraphics[trim=0cm 0.3cm 5.2cm 0.8cm , clip,width=0.95\linewidth]{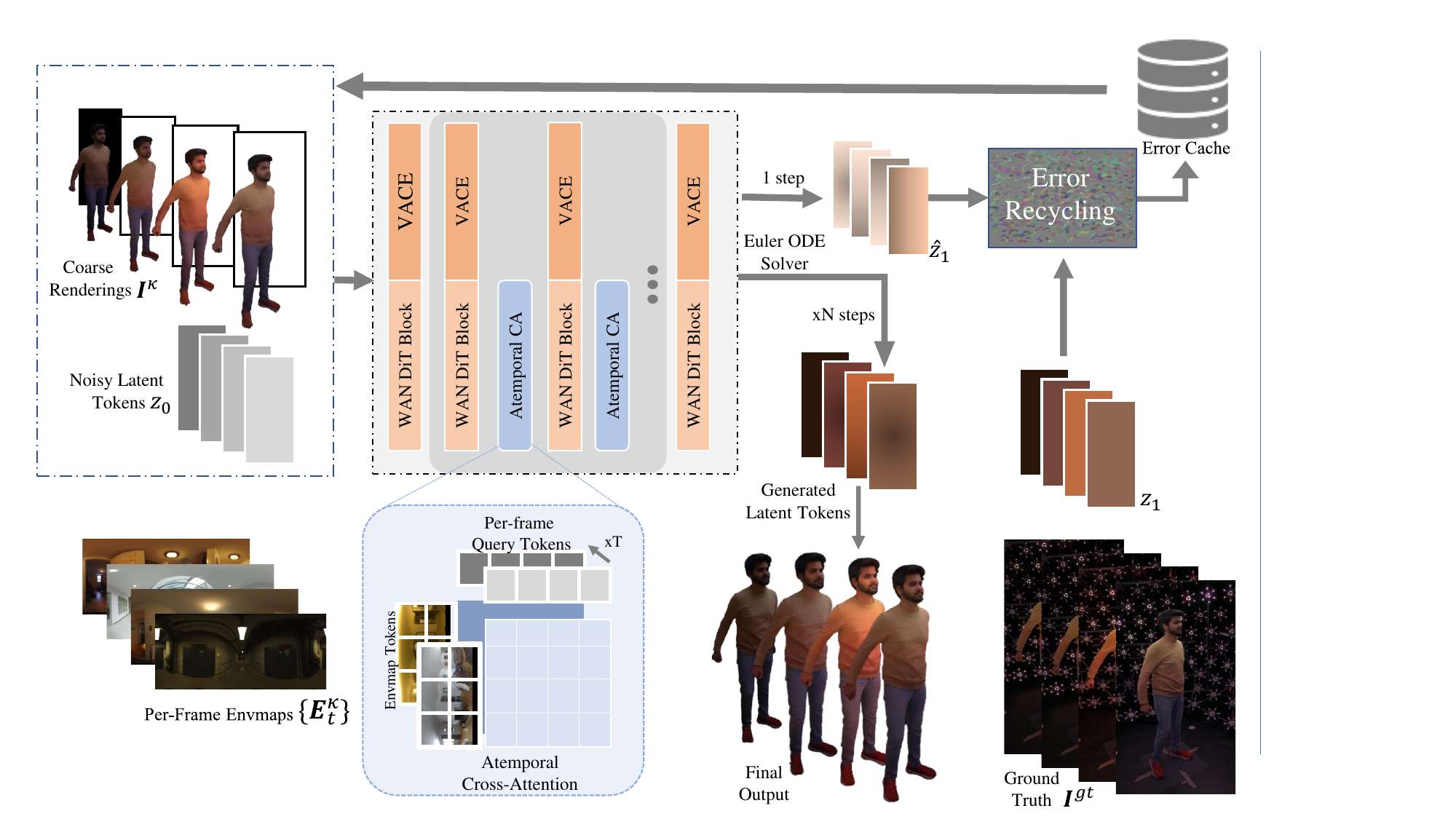}
    \caption{\textbf{Generative Refinement.} Given the coarse renderings of the character from a target viewpoint, we finetune the WAN2.1-VACE architecture to produce fine-grained renderings.
    Environment illumination is introduced through Atemporal Cross-Attention that performs per-frame attention between Envmap tokens and video tokens.}
     \vspace{-12pt}
    \label{fig:video}
\end{figure}

RelightNet follows the UNet-based architecture proposed in RHC~\cite{anon2025rhc} and operates entirely in the UV texture space.
It uses cross-attention to condition the texture prediction on the environment map $\mathbf{E}$, enabling efficient relighting of dynamic human characters under arbitrary illumination. The predicted texture is trained using L1 loss with ground truth images.
Unlike RHC~\cite{anon2025rhc}, which leverages image observations to transfer details in texture space, we rely only on the coarse geometry and derived features, recovering fine details generatively (Sec.~\ref{sec:gen_refine}).
\vspace{-12pt}
\subsection{Generative Refinement} \label{sec:gen_refine}
While RelightNet produces physically grounded appearance, its deterministic output $\boldsymbol{I}_t^{\boldsymbol{\kappa}}$ tends to regress to the mean for stochastic high-frequency details such as dynamic wrinkles.
To recover these details, we formulate a probabilistic solution by introducing a generative refinement stage that treats the final appearance as a conditional distribution. 
We leverage a pretrained video generation backbone based on \textit{Latent Flow Matching (LFM)}~\cite{lipman2022flow}, adapted to incorporate lighting controls. 
Importantly, we do so in the image space instead of the UV texture space to best leverage the learned priors of the pre-trained video models.

We operate in the latent space $\mathcal{Z}$ of a pre-trained 3D VAE, where a video sequence $\mathcal{V}$ is encoded as latent $\boldsymbol{z} = \mathcal{E}(\mathcal{V})$ via the VAE encoder. LFM defines a velocity field $v$ that describes the probability path $p_s(\boldsymbol{z})$ between an initial noise distribution $p_0 = \mathcal{N}(0, \mathbf{I})$ at flow-step $s=0$ and the target data distribution $p_1$ at $s=1$. 
Following LFM~\cite{lipman2022flow}, we define a linear probability path:
$ \boldsymbol{z}_s = (1-s)\boldsymbol{z}_0 + s\boldsymbol{z}_1$, where $\boldsymbol{z}_1$ is the encoded clean latent and $\boldsymbol{z}_0 \sim p_0$ is Gaussian noise. 
The velocity field $v(\boldsymbol{z}_s, s, \mathbf{c})$ is modeled by a neural network trained to regress the field by minimizing: 
\begin{equation} \label{eq:fm_loss}
    \mathcal{L}_{\text{FM}} = \mathbb{E}_{s} \left\| v(\boldsymbol{z}_s, s, \mathbf{c}) - (\boldsymbol{z}_1 - \boldsymbol{z}_0) \right\|^2,
\end{equation}
where $s \sim \mathcal{U}[0,1]$ and $\mathbf{c}$ is the conditional context. During inference, we solve the Ordinary Differential Equation (ODE) from $s=0$ to $s=1$ %
using an ODE solver (e.g., Euler). The final conditionally generated video is recovered via the VAE decoder $\hat{\mathcal{V}} = \mathcal{D}(\boldsymbol{z}_0)$. 

In our case, the conditioning set $\mathbf{c} = \{\mathbf{I}^{\boldsymbol{\kappa}}, \mathbf{E}\}$ comprises of the coarse image-space renderings of the RelightNet output and the target illumination.
Since the coarse renderings $\mathbf{I}^{\boldsymbol{\kappa}}$ are pixel-aligned with the target photorealistic output, this can be formulated as a video-to-video generation task, allowing us to use WAN2.1-VACE~\cite{jiang2025vace} as the pretrained backbone. 
Built on top of WAN2.1~\cite{wan2025wanopenadvancedlargescale}, WAN2.1-VACE is a video-to-video model that allows video generation with spatio-temporal control derived from a given reference video.
In addition, we propose to use atemporal cross-attention to introduce the per-frame environment maps $\mathbf{E} \in \mathbb{R}^{T\times256\times512\times3}$ into the WAN2.1 architecture.
\vspace{-12pt}
\subsubsection{VACE Control.} \label{sec:gen_vace}

To introduce precise spatio-temporal control from the RelightNet output, the WAN DiT backbone is augmented using VACE~\cite{jiang2025vace}, which extends the base model with adapter layers that accept a reference video as a spatio-temporal conditioning signal.

Central to VACE is a \emph{concept decoupling} mechanism that separates the conditioning signal into two complementary components via a Video Control Unit (VCU). Given the coarse RelightNet rendering $\mathbf{I}^{\boldsymbol{\kappa}} \in \mathbb{R}^{T \times H \times W \times 3}$ and a per-frame binary mask $\mathbb{M} \in \mathbb{R}^T$, frames are partitioned into \emph{reactive} frames, which the model actively refines, and \emph{inactive} frames, which serve as fixed temporal anchors:
\begin{equation}
    F_{\mathrm{reac}} = \mathbf{I}^{\boldsymbol{\kappa}} \odot \mathbb{M}, 
    \quad 
    F_{\mathrm{inac}} = \mathbf{I}^{\boldsymbol{\kappa}} \odot (1 - \mathbb{M}).
\end{equation}
The reactive component captures the frames to be refined under the target illumination, while the inactive component provides stable spatial and temporal context that anchors the generation and prevents content drift. Both components, 
together with the mask $\mathbb{M}$, are encoded via the 3DVAE encoder and concatenated channel-wise to form the VCU input. This is processed by dedicated attention blocks within the VACE adapter and fused into each DiT block of the 
WAN2.1 backbone as an additive residual, propagating the spatio-temporal conditioning signal throughout the full depth of the network.

In addition to the coarse renderings, we provide a reference image $\mathbf{I}^{\mathrm{ref}}$ of the subject captured under neutral illumination from the training set, which is passed to the VACE VCU alongside the coarse renderings. We append one additional token to the DiT corresponding to this reference image, with a velocity target defined from noise to its encoded latent. Since the model must reproduce the reference image at this token, it is encouraged to attend to the subject's identity-specific appearance independently of the target lighting condition.

During training, we leverage this decoupling to teach the model temporal continuation. Specifically, we replace the first $5$ frames of $\mathbf{I}^{\boldsymbol{\kappa}}$ with ground-truth frames with probability $40\%$, marking them as inactive with $\mathbb{M}_{1:5} = 0$ and setting $\mathbb{M}_{6:T} = 1$ for the remaining frames. This trains the model to treat the seed frames as a fixed spatio-temporal reference and coherently continue the sequence from them, which is the key capability exploited during infinite rollout at inference time (Sec.~\ref{sec:gen_error}).
\vspace{-12pt}
\subsubsection{Atemporal Lighting Conditioning.} \label{sec:gen_atemp_env}

We train our model on the RHC dataset~\cite{anon2025rhc}, which captures human performance under a distinct environment map in each frame. While the coarse RelightNet renderings $\mathbf{I}^{\boldsymbol{\kappa}}$ encode approximate lighting, they lack explicit global illumination context, forcing the generative backbone to infer target lighting from limited cues.

To provide direct per-frame lighting information, we introduce \emph{atemporal lighting conditioning}, which encodes each environment map $\{\mathbf{E}_t\}$ (transformed to camera space $\{\boldsymbol{E}_t^{\kappa}\}$) into a latent space using the WAN2.1 3DVAE encoder:
\begin{equation}
    \boldsymbol{\ell} = \mathcal{E}(\{\boldsymbol{E}_t^{\kappa}\}),
\end{equation}
yielding temporally-aligned embeddings $\boldsymbol{\ell} \in \mathbb{R}^{T^{\prime} \times H^{\prime} \times W^{\prime} \times d}$. Each latent $\boldsymbol{\ell}_i$ is injected into the corresponding DiT feature $\boldsymbol{f}_i$ via cross-attention,
with $\tilde{\boldsymbol{f}} = \operatorname{Concat}_i(\mathrm{CrossAttn}(\boldsymbol{f}_i, \boldsymbol{\ell}_i))$. No temporal cross-attention is applied, ensuring each frame is conditioned only on its own environment map.

As shown in the OLAT results of our supplementary video, this explicit per-frame illumination context enables the model to faithfully reproduce fine-grained, temporally varying lighting while preserving motion priors.
\vspace{-12pt}
\subsubsection{Error Recycling.} \label{sec:gen_error}

Although history-anchored conditioning (Sec.~\ref{sec:gen_vace}) improves temporal continuation, long-horizon generation remains susceptible to drift due to the distribution mismatch between ground-truth context observed during training and self-generated context at inference.

To mitigate this, we adopt \emph{Error Recycling}, recently introduced in Stable Video Infinity~\cite{li2025stable}. The key idea is to re-inject reconstruction residuals from previous model predictions as stochastic perturbations during subsequent training iterations. This explicitly trains the model to recover from errors induced by its own predictions. Concretely, given potentially perturbed clean and noisy latents $\tilde{\boldsymbol{z}}_1$ and $\tilde{\boldsymbol{z}}_0$, we construct the noisy latent
$
\tilde{\boldsymbol{z}}_s = (1-s)\tilde{\boldsymbol{z}}_0 + s\tilde{\boldsymbol{z}}_1,
$
and define the target velocity
$
v_s^{\text{rcy}} = \boldsymbol{z}_1 - \tilde{\boldsymbol{z}}_0,
$
where the target always points to the clean latent $\boldsymbol{z}_1$ regardless 
of the perturbation applied to $\tilde{\boldsymbol{z}}_0$, encouraging the model 
to correct corrupted trajectories back toward clean outputs.

The model predicts velocity $v(\tilde{\boldsymbol{z}}_s, s, \mathbf{c})$, from which we obtain first-order estimates of the clean and noisy latents:
\vspace{-4pt}
\begin{equation}
\vspace{-6pt}
\hat{\boldsymbol{z}}_0 = \tilde{\boldsymbol{z}}_s - s\, v(\tilde{\boldsymbol{z}}_s, s, \mathbf{c}), 
\quad
\hat{\boldsymbol{z}}_1 = \tilde{\boldsymbol{z}}_s + (1-s)\, v(\tilde{\boldsymbol{z}}_s, s, \mathbf{c}).
\end{equation}

\begin{figure*}[t]
\centering
\setlength{\tabcolsep}{0pt}
\renewcommand{\arraystretch}{0}

\begin{tabular}{ccc}
\includegraphics[
    page=4,
    width=0.333\textwidth,
    height=0.333\textwidth,
    trim=210pts 0pts 210pts 0pts,
    clip
]{imgs/gra_qual.pdf}
&
\includegraphics[
    page=6,
    width=0.333\textwidth,
    height=0.333\textwidth,
    trim=210pts 2pts 210pts 2pts,
    clip
]{imgs/gra_qual.pdf}
&
\includegraphics[
    page=5,
    width=0.333\textwidth,
    height=0.333\textwidth,
    trim=210pts 2pts 210pts 2pts,
    clip
]{imgs/gra_qual.pdf}
\\[0pt]

\includegraphics[
    page=2,
    width=0.333\textwidth,
    height=0.333\textwidth,
    trim=210pts 2pts 210pts 2pts,
    clip
]{imgs/gra_qual.pdf}
&
\includegraphics[
    page=3,
    width=0.333\textwidth,
    height=0.333\textwidth,
    trim=210pts 2pts 210pts 2pts,
    clip
]{imgs/gra_qual.pdf}
&
\includegraphics[
    page=1,
    width=0.333\textwidth,
    height=0.333\textwidth,
    trim=210pts 2pts 210pts 2pts,
    clip
]{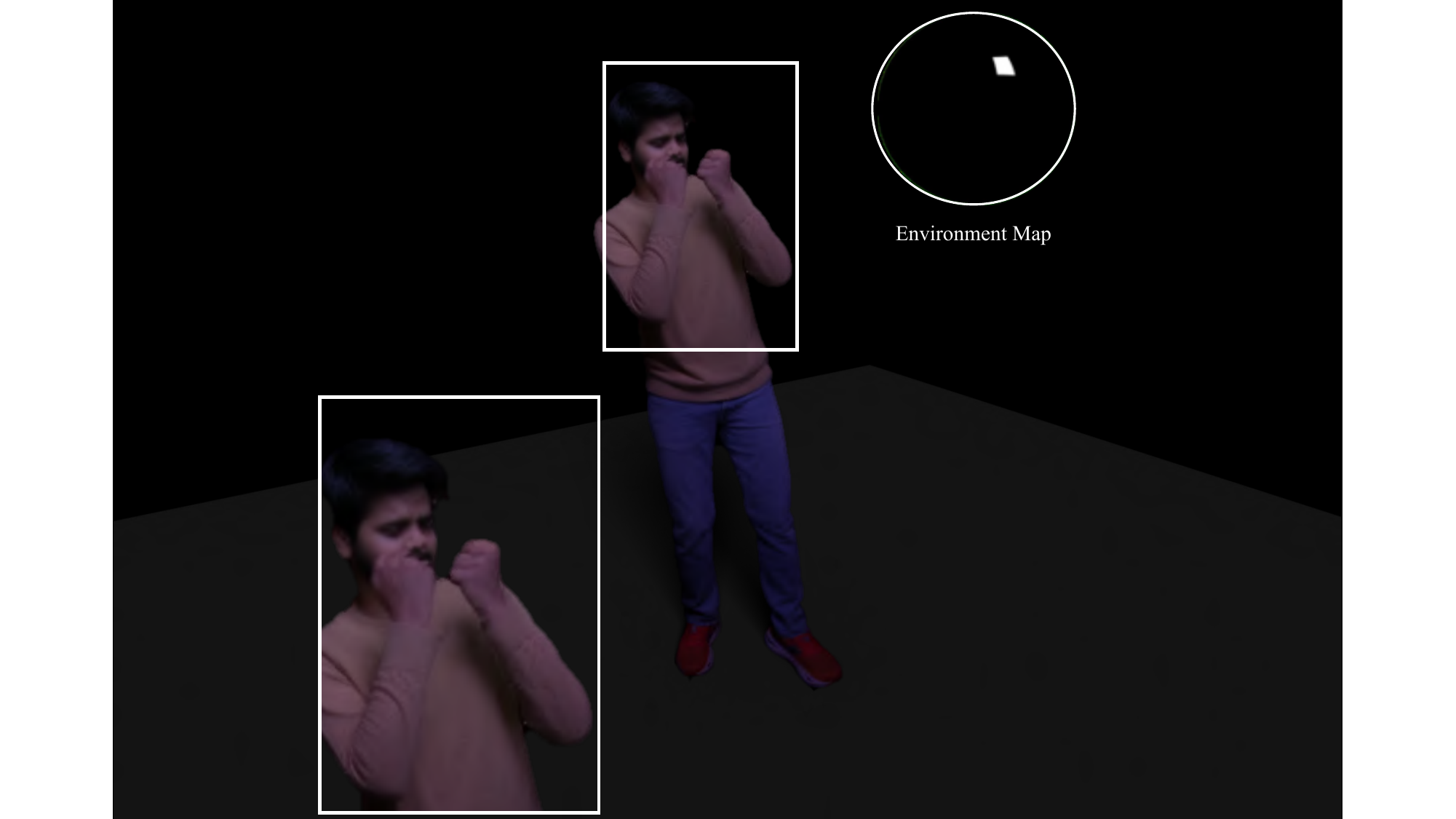}
\end{tabular}

\caption{\textbf{Qualitative Results.} GRA generates photorealistic free-view renderings of dynamic humans under novel poses, viewpoints, and illumination. Characters are seamlessly placed into diverse real-world environments, demonstrating realistic relighting, coherent shading, and consistent appearance under challenging lighting conditions. Notably, our physics-grounded model is able to generalize to OLAT lighting conditions despite not having seen these at training time.}
\vspace{-18pt}
\label{fig:qual}
\end{figure*}

Analogously, using the target velocity $v_s^{\text{rcy}}$, we define reference reconstructions to obtain a consistent reference under the ground-truth velocity:
\vspace{-4pt}
\begin{equation}
\vspace{-6pt}
\boldsymbol{z}_0^{\text{rcy}} = \tilde{\boldsymbol{z}}_s - s\, v_s^{\text{rcy}}, 
\quad
\boldsymbol{z}_1^{\text{rcy}} = \tilde{\boldsymbol{z}}_s + (1-s)\, v_s^{\text{rcy}}.
\end{equation}

The reconstruction residuals can then be computed as:
$
\boldsymbol{e}_0 = \hat{\boldsymbol{z}}_0 - \boldsymbol{z}_0^{\text{rcy}} $, and 
$
\boldsymbol{e}_1 = \hat{\boldsymbol{z}}_1 - \boldsymbol{z}_1^{\text{rcy}}$.
These residuals are stored in an \emph{Error Cache} and reused in subsequent iterations to perturb all latents involved in history-conditioned training:
\vspace{-6pt}
\begin{equation}
\vspace{-2pt}
\tilde{\boldsymbol{z}}_0 \leftarrow \boldsymbol{z}_0 + \boldsymbol{e}_0, \quad
\tilde{\boldsymbol{z}}_1 \leftarrow \boldsymbol{z}_1 + \boldsymbol{e}_1, \quad
\tilde{\boldsymbol{z}}_{\mathrm{inac}} \leftarrow \boldsymbol{z}_{\mathrm{inac}} + \boldsymbol{e}_1,
\end{equation}
where $\boldsymbol{z}_{\mathrm{inac}} = \mathcal{E}(F_{\mathrm{inac}})$ denotes the VAE-encoded inactive conditioning frames. Each perturbation is applied independently with probabilities $0.01$, $0.9$, and $0.9$, respectively. Training proceeds with the standard velocity regression objective under the perturbed latent distribution:
\begin{equation}
\mathcal{L}_{ER}
=
\mathbb{E}_{s}
\left\|
v(\tilde{\boldsymbol{z}}_s, s, \mathbf{c})
-
(\boldsymbol{z}_1 - \tilde{\boldsymbol{z}}_0)
\right\|^2.
\end{equation}

\noindent By exposing the model to trajectories corrupted by its own reconstruction residuals, error recycling explicitly regularizes against accumulation of temporal drift and improves stability in long-form generation. As demonstrated in the supplementary video, it plays a key role in preserving high-frequency motion fidelity over extended sequences. For training details, see the supplementary material.

\vspace{-12pt}
\section{Experiments} \label{sec:experiment}
\vspace{-12pt}
We demonstrate that our method produces photorealistic renderings under novel poses, viewpoints, and illumination conditions in Fig.~\ref{fig:qual}, across both indoor and outdoor environments. Owing to the physically grounded nature of our relighting formulation, our approach generalizes to one-light-at-a-time lighting setups even though such conditions are not observed during training.

\par \noindent \textbf{Dataset.}
We use the Relightable Holoported Characters (RHC) dataset, providing 40 synchronized camera viewpoints of five subjects performing diverse motions, with interleaved uniformly lit and relit frames. Training uses 1,015 HDR environment maps from the Laval Indoor dataset~\cite{gardner2017learning}; testing uses 8 unseen environments. Test sequences consist of 128 consecutive relit frames under a fixed environment. Each subject has 28,420 training and 14,336 testing frames per camera, captured at 60\,Hz. We evaluate on four subjects, training all components per-subject, with 37 cameras for training and 3 held out for testing.

\par \noindent \textbf{Metrics.}
We report PSNR, SSIM~\cite{wang2004image}, and LPIPS~\cite{zhang2018perceptual} averaged over every 10th frame across held-out test views, evaluated on the middle frame of a 17-frame generation window to avoid boundary effects. We also report FVD~\cite{liu2024fr}, computed on 17-frame clips sampled every 200 frames, to capture temporal coherence and overall video realism.

\begin{table*}[!t]
    \centering
    \caption{
        \textbf{Quantitative Evaluation.}
        We compare our method to prior methods for human performance relighting, including Relighting4D (R4D)~\cite{chen2022relighting4d}. 
        Following RHC~\cite{anon2025rhc}, we provide high-quality DDC-tracked meshes and calibrated ground-truth environment maps for multi-view training.
        This training protocol is further extended to IntrinsicAvatar (IA)~\cite{wang2024intrinsicavatar}, MeshAvatar (MA)~\cite{chen2024meshavatar}, and DNF-Avatar (DNFA)~\cite{jiang2025dnf}.
        Our method achieves the best perceptual quality across all subjects.
    }
    \vspace{-10pt}
    \label{tab:main}
    \setlength{\tabcolsep}{2.8pt} 
    \renewcommand{\arraystretch}{1.5} 
    \resizebox{1.0\textwidth}{!}{
    \begin{tabular}{|c?c | c | c | c ?c| c| c| c?c|c|c|c?c|c|c|c|}
        \hline
        \multirow{2}{*}{Method} & \multicolumn{4}{c?}{Subject 1 (S1)} & \multicolumn{4}{c?}{Subject 2 (S2)} & \multicolumn{4}{c?}{Subject 3 (S3)} & \multicolumn{4}{c|}{Subject 4 (S4)}   \\
        \cline{2-17}
         & PSNR$\uparrow$ & LPIPS$\downarrow$ & SSIM$\uparrow$ & FVD$\downarrow$ & PSNR$\uparrow$ & LPIPS$\downarrow$ & SSIM$\uparrow$ & FVD$\downarrow$ & PSNR$\uparrow$ & LPIPS$\downarrow$ & SSIM$\uparrow$ & FVD$\downarrow$ & PSNR$\uparrow$ & LPIPS$\downarrow$ & SSIM$\uparrow$ & FVD$\downarrow$ \\
         \hline
         R4D~\cite{chen2022relighting4d} & \uline{29.89} & 10.31 & \uline{87.15} & 247 & \uline{31.13} & 8.04 & \uline{87.08} & 161 & \textbf{29.32} & 10.62 & \textbf{87.67} & 186 & \uline{31.98} & 10.07 & \textbf{87.90} & 168 \\
         \hline
         IA~\cite{wang2024intrinsicavatar} & 27.25 & 18.25 & 81.39 & 408 & 28.87 & 15.43 & 82.07 & 345 & 26.14 & 22.52 & 79.07 & 621 & 29.50 & 18.46 & 82.91 & 354 \\
         \hline
         MA~\cite{chen2024meshavatar} & 29.71  & \uline{9.09}  & 84.69 & \uline{77} & 30.36  & \uline{8.03}  & 84.99 & \uline{74} & 28.35  & \uline{10.16}  & 84.60 & \uline{64} & 31.13  &  \uline{8.50} & 84.59 & \uline{73} \\
         \hline
         DNFA~\cite{jiang2025dnf} & 26.32 & 11.57 & 82.50 & 212 & 28.57 & 9.40 & 82.92 & 131 & 26.37 & 12.04 & 82.26 & 150 & 29.31 & 10.82 & 84.12 & 124 \\
         \hline
        \textbf{Ours} & \textbf{30.49} & \textbf{6.42} & \textbf{87.52} & \textbf{49} & \textbf{31.15} & \textbf{5.64} & \textbf{87.50} & \textbf{48} & \uline{29.11} & \textbf{7.40} & \uline{87.16} & \textbf{52} &  \textbf{32.09} & \textbf{6.55} & \uline{87.35} & \textbf{63} \\
        \hline 
    \end{tabular}}
    \vspace{-12pt}
\end{table*}

\par \noindent \textbf{Baselines.}
We compare against state-of-the-art relightable 3D avatar methods. For R4D~\cite{chen2022relighting4d}, we extend it to a motion-driven setting with multi-view supervision and ground-truth environment maps, replacing its SMPL mesh with our higher-quality DDC-tracked mesh for a fair comparison. We also compare against IntrinsicAvatar (IA)~\cite{wang2024intrinsicavatar}, MeshAvatar (MA)~\cite{chen2024meshavatar}, and DNF-Avatar (DNFA)~\cite{jiang2025dnf}, trained with multi-view supervision and ground-truth envmaps using their standard SMPL-based representations, as their tight coupling with SMPL/SMPL-X prevents substitution with our DDC mesh. We exclude Relightable Full-Body Gaussian Codec Avatars~\cite{wang2025relightable} as its implementation is not publicly available.

\vspace{-12pt}
\subsection{Comparisons on Novel Sequences, Viewpoints, and Lighting} 
\label{sec:exp_main}
\vspace{-8pt}
\textbf{Quantitative Comparisons.}
Tab.~\ref{tab:main} reports results on novel sequences under novel viewpoints and unseen lighting. Our method outperforms all baselines in perceptual quality across all subjects. R4D with DDC tracking surpasses IA and MA on PSNR and SSIM, confirming the importance of accurate geometry for pixel-wise metrics. MA
achieves sharper results than IA and R4D via its hybrid explicit-implicit representation, but remains limited by pose-only conditioning and restrictive BRDF assumptions. DNFA improves perceptual sharpness over its teacher IA via a more expressive Gaussian representation, but inherits its tracking limitations and remains less
robust than MA. Our method achieves the best perceptual and video performance across all subjects.

We note that PSNR and SSIM are imperfect proxies for relighting quality: both favor blurry, deterministic outputs, and fine-scale details like wrinkles are inherently ambiguous at test time. LPIPS and FVD better reflect the perceptual gains of our approach.

\begin{figure*}[t!]
    \centering
    \includegraphics[trim={8cm 3.4cm 11cm 0cm},clip,width=0.9\textwidth]{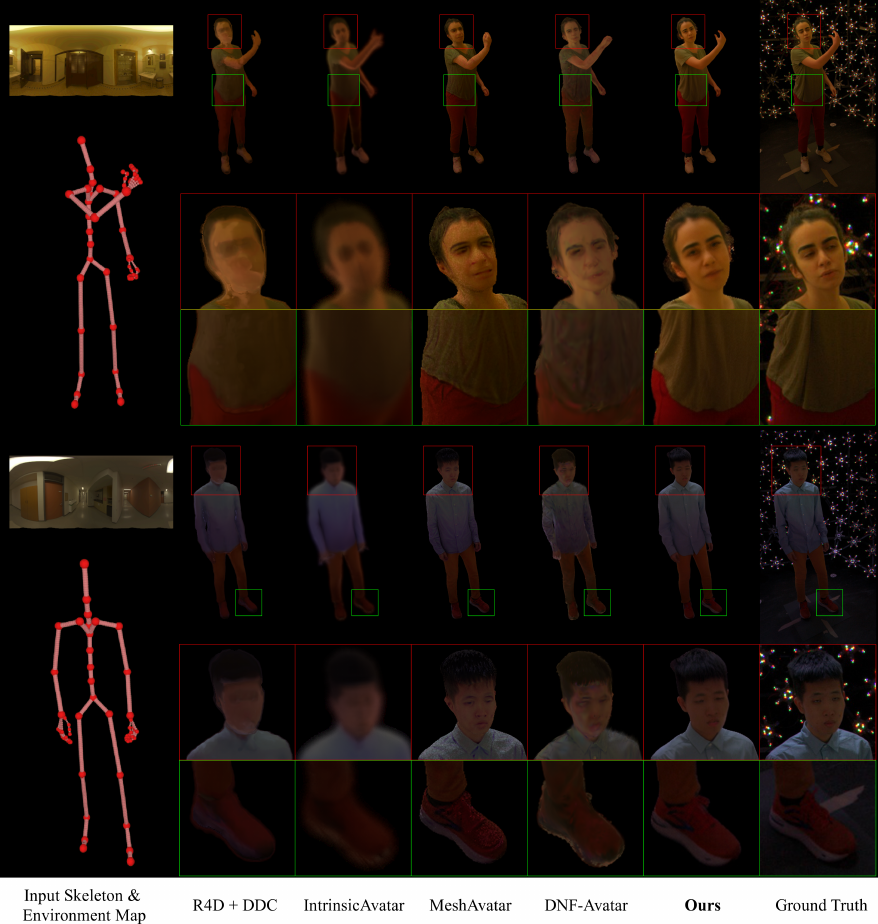}
    \caption{
    \textbf{Qualitative Comparison.}
    We compare our method with state-of-the-art relightable avatar methods: Relighting4D (R4D)~\cite{chen2022relighting4d} augmented with a DDC-tracked template, IntrinsicAvatar~\cite{wang2024intrinsicavatar}, MeshAvatar~\cite{chen2024meshavatar}, and DNF-Avatar~\cite{jiang2025dnf}. Our approach consistently improves visual realism and relighting quality.
    }
    \vspace{-21pt}
    \label{fig:exp-main}
\end{figure*}

\noindent \textbf{Qualitative Comparisons.} Fig.~\ref{fig:exp-main} shows our qualitative comparisons. Our method produces sharp, physically plausible relighting with dynamic wrinkles, while all baselines yield noticeably blurrier results due to deterministic pose conditioning. R4D produces over-smooth results. IA additionally suffers from SMPL tracking inaccuracies, leading to further loss of detail. MA generates comparatively sharper textures but fails to reproduce fine-scale material responses. DNFA is sharper than IA but exhibits visible high-frequency artifacts. Additional results are in the supplementary material.
\vspace{-12pt}
\subsection{Ablation Studies} \label{sec:exp_ablation}
\begin{table}[t]
    \footnotesize 
    \centering
    \caption{
    \textbf{Ablation Study.}
    We ablate different design choices of our method. 
    Note the progressive improvement by the addition of individual components. When removing individual components from the pipeline, the results degrade confirming the importance of our individual design choices. 
    \vspace{-6pt}
    }
    \label{tab:ablation}
    \begin{tabular}{|l|c|c|c|c|}
         \hline
         Method & PSNR $\uparrow$ & LPIPS  $\downarrow$ &SSIM $\uparrow$ &FVD $\downarrow$\\
         \hline
         Microfacet Rendering & 29.34  & 12.26 & 86.85& 178.42\\
          \quad + RelightNet & \textbf{30.90} & 8.92 & \textbf{89.25} & 118.24 \\
          \qquad + \textbf{Generative Refinement (Ours)}  & 30.49 & \textbf{6.42} & 87.52& \textbf{48.79} \\
          \hline
         Ours w/ Temporal Lighting Conditioning & \uline{30.78} & 6.53 & \uline{88.00}& \uline{58.18} \\
         Ours w/o Atemporal Lighting Conditioning & 30.53 & \uline{6.48} & 87.76 & 58.21 \\
         Ours w/o VACE & 30.35 & 6.93 & 87.12 & 61.84 \\
         Ours w/o RelightNet & 29.56 & 6.66 & 87.26 &63.81 \\
         Ours w/o VACE Finetuning  & 24.92 & 22.43 & 70.76 & 230.94\\
         Ours w/o Prior & 30.02 & 9.60 & 86.67 & 394.91\\
         Ours w/ Neural Gaffer~\cite{jin2024neural} & 29.33 & 9.25 & 86.62 & 102.10 \\
         Ours w/ Few-shot Adaptation & 29.82 & 7.01 & 87.30 & 65.25 \\
         \hline
    \end{tabular}
    \vspace{-12pt}
\end{table}

We conduct quantitative (Tab.~\ref{tab:ablation}) and qualitative (Suppl. Mat.) ablation studies to validate each component. Full details are provided in the supplementary material. The microfacet rendering achieves reasonable PSNR from physically correct low-frequency shading but lacks fine-scale detail. RelightNet improves PSNR and SSIM via learned pose-correlated appearance, but its determinism limits perceptual fidelity, motivating the generative refinement stage. Atemporal cross-attention outperforms temporal cross-attention by inducing per-frame lighting correspondence as an inductive bias rather than requiring the model to learn it implicitly. Without VACE finetuning, the model fails to treat coarse renderings as a shading target, producing physically inconsistent results. Training without the pretrained video prior fails to converge within the same budget, and replacing our video model with an image-based alternative (Neural Gaffer~\cite{jin2024neural}) yields severe temporal inconsistency (as shown by the degraded FVD), confirming the necessity of video-level temporal modeling.

\subsection{Discussion} \label{sec:discussion}
We now discuss the applications and broader implications of our design choices.

\noindent \textbf{Physics-grounded generalization.}
Our physics-grounded formulation enables generalization to lighting conditions not seen during training. We demonstrate strong generalization to one-light-at-a-time (OLAT) environment maps (Fig.~\ref{fig:teaser}, Fig.~\ref{fig:qual}, and Suppl.\ Vid.), where the model produces stable, physically plausible relighting under sparse, highly directional illumination despite being trained primarily on natural indoor environments.
This generalization stems from the explicit microfacet rendering, which provides a physically consistent inductive bias that guides the generative model toward correct shading even under unseen distributions.

\begin{figure*}[t!]
\centering
\setlength{\tabcolsep}{0pt}
\renewcommand{\arraystretch}{0}

\begin{tabular}{cc}
\centering
    
    \includegraphics[page = 1, trim = 4.7cm 0.0cm 1cm 0cm, clip, width=0.496\textwidth]{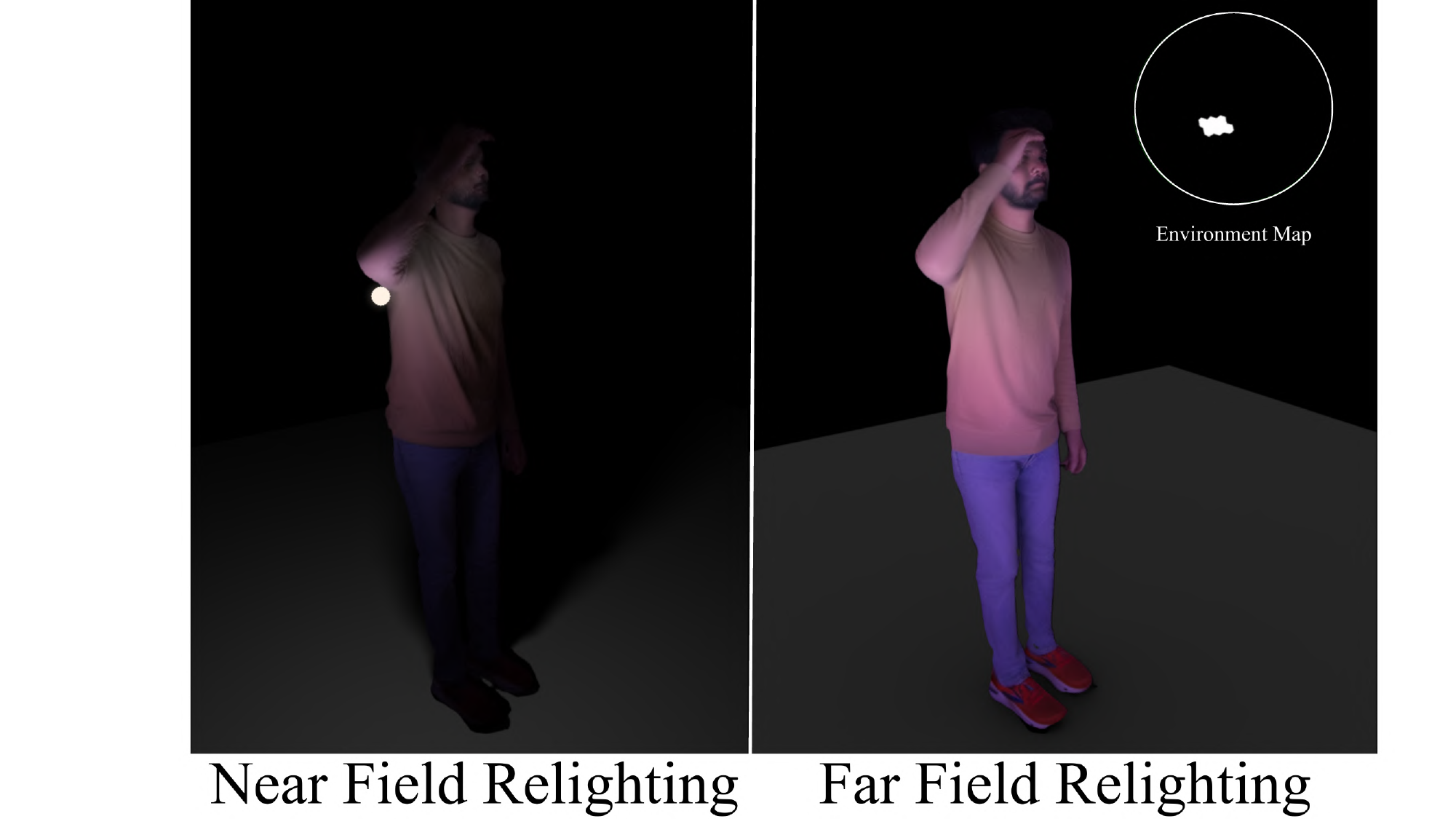} & 
    \includegraphics[page = 2, trim = 1.4cm 0.0cm 3.9cm 0.0cm, clip, width=0.505\textwidth]{imgs/gra_appl101.pdf} 
\end{tabular}
    \vspace{-8pt}
    \caption{\textbf{Physics Grounded Applications.} We demonstrate two capabilities of our \emph{w/o RelightNet} model enabled by its explicit physics-grounded design. (Left) A local near-field point light source produces physically plausible shading, including arm self-shadowing. For comparison, our full model approximates the same light as far-field illumination by projecting it onto the lightstage. (Right) Relightable texture editing is achieved by directly modifying the albedo map in UV space.} %
     \vspace{-21pt}
    \label{fig:nfield}
\end{figure*}

\noindent \textbf{Near-field relighting.}
Interestingly, our method also supports near-field relighting (Fig.~\ref{fig:teaser}, Fig.~\ref{fig:nfield}, and Suppl.\ Vid.) via the \emph{w/o RelightNet} variant. We render microfacet-based shading under a point light source, and pass the resulting coarse renderings directly to the video model (Suppl. for details). %
The fact that this approximation produces plausible results demonstrates that the physics-based design, combined with the generative prior, is expressive enough to handle localized, unseen lighting configurations without
any dedicated training.

\noindent \textbf{Few-shot identity adaptation.}
The subject-specific finetuning required 
\begin{wrapfigure}[15]{r}{0.5\columnwidth}
    \includegraphics[page=1,trim={5.5cm 8.5cm 12cm 3cm},clip,width=0.95\linewidth]{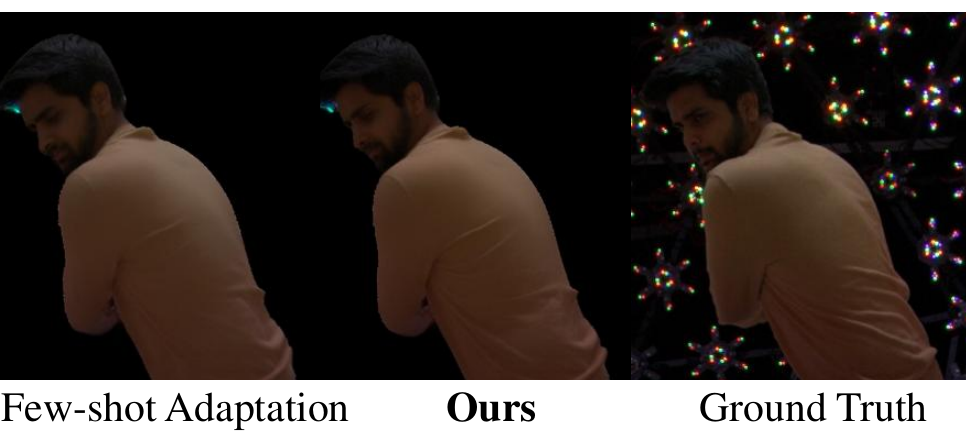}
    \vspace{-9pt}
    \caption{Starting from a Subject~4 checkpoint, we finetune our video model on $170$ frames of Subject~1. The adapted model produces photorealistic, temporally consistent relighting for the new identity despite minimal supervision.} 
    \label{fig:gen}
\end{wrapfigure}

by our pipeline raises a natural question about generalization across identities. 
Our few-shot adaptation experiment (Fig.~\ref{fig:gen}) shows that despite prior subject-specific adaptation, the video model retains sufficient generative capacity to transfer to a new identity from as little as 6 seconds of capture data. This suggests that subject-specific finetuning primarily anchors identity-specific appearance without catastrophically forgetting the broader motion and illumination priors learned during pretraining.

\noindent \textbf{Relightable texture editing,}
as shown in 
(Fig.~\ref{fig:nfield}), is similarly enabled by the explicit material decomposition of the \emph{w/o RelightNet} model. Modifying the optimized UV albedo map and re-rendering with the microfacet model produces edited coarse renderings that the video model refines into photorealistic output. The generative stage faithfully preserves the intended texture modifications while synthesizing consistent shading, demonstrating that explicit material control naturally extends to appearance editing with no additional supervision.

\vspace{-12pt}

\vspace{-8pt}
\section{Conclusion}
\vspace{-12pt}
In this paper, we proposed GRA as a hybrid method to build generative 3D relightable avatars.
We showed that integrating explicit, physics-grounded texture appearance with video generative models allows us to have 3D control as well as photorealism at the same time.
Interestingly, we showed that our approach can also generalize to out-of-distribution target illuminations, such as OLAT illumination without any retraining.
As it is grounded in explicit 3D representation, GRA can model global illumination effects like self-shadows.
Finally, we demonstrated the ability of our system to generate long-form videos, a capability not easily afforded by video generative models.
In the future, we intend to model the face and hands in high-resolution, while also introducing near-field relighting effects within this framework.
\vspace{-12pt}
\section{Acknowledgements}
\vspace{-6pt}
This work was funded by the Deutsche Forschungsgemeinschaft (DFG, German Research Foundation) -- GRK 2853/1 “Neuroexplicit Models of Language, Vision, and Action” - project number 471607914.

\bibliographystyle{splncs04}
\bibliography{main}
\clearpage

\appendix
\section{Overview}
This supplementary document provides additional details and analyses to complement the main paper. 
We begin with a description of the RHC~\cite{anon2025rhc} dataset (Sec.~\ref{sec:s-dataset}), followed by a detailed exposition of our method (Sec.~\ref{sec:s-method}). 
Training and implementation details are provided in Sec.~\ref{sec:s-training}. Baseline implementations are discussed in Sec.~\ref{sec:s-baseline}, and runtime details in Sec.~\ref{sec:s-runtime}. 
Sec.~\ref{sec:s-ablations} presents the ablation studies referenced in the main paper, while Sec.~\ref{sec:s-nfield} describes the implementation of our near-field relighting experiment. 
Details of the supplementary video, including content and visualizations, are given in Sec.~\ref{sec:s-video}. Comparisons to recent Gaussian-based relighting methods are provided in Sec.~\ref{sec:s-rhc}, and additional qualitative results are presented in Sec.~\ref{sec:s-qual}. 
Finally, Sec.~\ref{sec:s-limits} discusses the limitations of our approach and outlines directions for future work.
\section{Details about the Dataset}\label{sec:s-dataset}

The RHC dataset~\cite{anon2025rhc} contains frames captured under interleaved lighting conditions, alternating between uniformly lit frames and frames illuminated by randomly sampled environment maps projected onto the light stage. The uniformly lit frames enable robust tracking, while the relit frames simulate diverse real-world lighting conditions.

The dataset includes 5 subjects with 40 viewpoints of 28,420 training frames and 14,336 testing frames at a resolution of $4096\times2160$, which we downsample to $1024\times540$ for all subsequent processing. The training sequence contains 1,015 unique lighting conditions sampled from the Laval Indoor~\cite{gardner2017learning} dataset. The testing sequences contain 8 held-out lighting conditions drawn from the same dataset. These lighting conditions have been normalized to match the maximum intensity of the lights in the lightstage.

Training and testing sequences were captured separately to ensure variation in the performed motions. Additionally, we hold out three cameras from the training sequences for novel-view synthesis evaluation.

\section{Method Details} \label{sec:s-method}
\subsection{Details about Microfacet Rendering}\label{sec:s-micro} 
We compute microfacet rendering in the 512x512 UV-space of the tracked template mesh. For each texel, we compute per light visiblity using Optix~\cite{parker2010optix} ray tracer. We first map the light sources from the environment map to the light emitters in the lightstage with pre-calibrated 3D position of LEDs. Then, we use this position to calculate the direction of incoming light for the BRDF calculation as well as to compute visibility. 

\subsection{Details about RelightNet}\label{sec:s-relightnet}
RelightNet consists of interleaved convolutional, self-attention and cross-attention layers for its encoder and decoder. 
Due to the high computational cost of self-attention and cross-attention at high resolution ($\ge 64\times64$), we only use convolutional layers at these resolutions.
For higher layer of the \textit{RelightNet} with low resolution feature map, we follow every convolutional layer with a self-attention layer and a cross-attention layer between the UV space features and the environment map. 
Here, self-attention is computed between flattened elements of the feature map.
The number of heads in self-attention and cross-attention is set as $4$.
Table~\ref{tab:s-na} illustrates the concrete architecture of \textit{RelightNet}. 
\begin{table}[h!]
    \centering
    \caption{Illustration of the \textit{RelightNet} architecture. In the operation column, "C" denotes convolution layer, "SA" denotes self-attention layer, "CA" denotes cross-attention layer, "DS" and "US" denotes down-sampling and up-sampling layer with scale factors equal to 2.}
    \begin{tabular}{|c|c|c|}
        \hline
         & Number of Feature Channels & Operation  \\
        \hline
         Input & 15 & N/A \\
         \hline
         Block 1 & (32,32,32) & (C, C, DS) \\ 
         Block 2 & (48,48) & (C, DS) \\ 
         Block 3 & (64,64) & (C, DS) \\ 
        \hline
         Block 4 & (64,64,64,64) & (C, SA, CA, DS) \\ 
         Block 5 & (128,128,128,128) & (C, SA, CA, DS) \\ 
         Block 6 & (256,256,256,256) & (C, SA, CA, DS) \\ 
         Block 7 & (256,256,256) & (C, SA, CA) \\ 
         Block 8 & (256,256,256,256) & (C, SA, CA, C) \\ 
         Block 9 & (256,256,256) & (C, SA, CA) \\ 
         Block 10 & (256,256,256,256) & (C, SA, CA, US) \\ 
         Block 11 & (256,256,256) & (C, SA, CA) \\ 
         Block 12 & (256,256,256, 256) & (C, SA, CA, US) \\ 
         Block 13 & (128,128,128) & (C, SA, CA) \\ 
         Block 14 & (128,128,128,128) & (C, SA, CA, US) \\ 
         Block 15 & (64,64,64) & (C, SA, CA) \\ 
         \hline
         Block 16 & (64,64) & (US, C) \\ 
         Block 17 & (48,48) & (US, C) \\ 
         Block 18 & (32,32,14) & (US, C, C) \\ 
         \hline
         Output & 3 & N/A \\
         \hline
    \end{tabular}
    \label{tab:s-na}
\end{table}

\section{Training and Implementation Details} \label{sec:s-training}

\noindent \textbf{Character Animation Module.}
We train the Deep Dynamic Characters (DDC) model using multi-view sequences captured under uniform lighting, along with Neus2~\cite{wang2023neus2} reconstructions for 3D supervision provided by the RHC dataset. We follow the training protocol of~\cite{habermann2021real}. This stage produces a temporally coherent, tracked template mesh $\boldsymbol{M}_t$ for each subject, which is used by all subsequent components.

\noindent \textbf{Microfacet Parameters.}
The per-texel albedo and roughness maps, denoted by $\boldsymbol{a}$ and $\boldsymbol{r}$, are optimized under a single-bounce microfacet rendering assumption.
Given a tracked mesh pose $\boldsymbol{M}_t$, camera $\boldsymbol{\kappa}$, and environment illumination $\boldsymbol{E}_t$, we obtain a rendered image via differentiable rendering $\mathcal{P}(\cdot)$.
We minimize an $\ell_1$ reconstruction loss against multi-view ground-truth relit images $\boldsymbol{I}_{t, \boldsymbol{\kappa}}^{\mathrm{gt}}$:

\begin{equation}
\mathcal{L}_{\mathrm{mf}} = \sum_{\boldsymbol{\kappa}, t} \left\| \mathcal{P}\!\left(M_t, \boldsymbol{\kappa}, \mathbf{E};
\boldsymbol{a}, \boldsymbol{r}\right) - \boldsymbol{I}_{t,\boldsymbol{\kappa}}^{\mathrm{gt}} \right\|_1 ,
\end{equation}
Optimization is performed using Adam with a learning rate of $10^{-5}$ for 13K iterations on a single A40 GPU, using batch size 1 and 8 gradient accumulation steps.

\noindent \textbf{RelightNet.}
RelightNet $\mathcal{R}$ is trained to refine the coarse physics-based renderings by modeling pose-dependent appearance variations and learned global illumination effects.
Training uses the same multi-view $\ell_1$ reconstruction loss as above.
We optimize $\mathcal{R}$ using Adam with a learning rate of $10^{-4}$ for 50K iterations on 4 A40 GPUs, with batch size 1 per GPU and 2 gradient accumulation steps.

\noindent \textbf{WAN2.1-VACE Finetuning.}
We initialize from a pretrained WAN2.1-VACE model and finetune it using multi-view image sequences.
We apply LoRA~\cite{hu2021lora} adaptation to the DiT backbone with rank 32, fully finetune the VACE adaptor, and train the atemporal lighting conditioning layers from scratch.
The model is trained using the flow-matching objective (Eq.~11 in the main paper), jointly conditioning on RelightNet outputs and per-frame environment map latents (extracted from the 3DVAE of WAN2.1).
Training is performed on 17-frame clips at a resolution of $1024\times544$. Since the original images are $1024\times540$, we apply zero-padding to match the required resolution.

\noindent \textbf{Data Augmentation for Generative Refinement.}
Our dataset is biased toward static cameras with a consistently moving subject. To increase scenario diversity, we apply several synthetic augmentations. With a probability of 10\%, we construct a clip by repeating a single frame 17 times. For such static clips, we simulate camera motion (pan/zoom) with a probability of 90\%. The motion is implemented as image-space translation and scaling using bicubic interpolation, applied consistently to both the input and ground-truth frames. For dynamic clips, simulated camera motion is applied with a probability of 5\% using the same transformation.
Training is performed for 4k iterations on four H100 GPUs with a batch size of 1 per GPU and two gradient accumulation steps.

\noindent \textbf{Atemporal Lighting Conditioning.}
For the Atemporal Lighting Conditioning, the environment maps are passed through the WAN-3DVAE and compressed into latent space. Now, every token of this latent space is passed through a shallow, shared MLP before being integrated into the WAN DiT using Atemporal Cross Attention. 

\section{Baseline Implementation Details} \label{sec:s-baseline}

We follow the training and evaluation protocols of~\cite{anon2025rhc} for Relighting4D~\cite{chen2022relighting4d} and IntrinsicAvatar~\cite{wang2024intrinsicavatar}.

For MeshAvatar~\cite{chen2024meshavatar}, we deviate from Singh et al.~\cite{anon2025rhc} by using SMPL-X tracking instead of SMPL tracking, as provided by the RHC dataset, while keeping all other settings unchanged.

For DNF-Avatar~\cite{jiang2025dnf}, we extend the training to multi-view supervision and provide ground-truth annotated environment maps, replacing the optimized environment maps used in the original method.

\section{Runtime Details}
\label{sec:s-runtime}
We report H100 runtime in Tab.~\ref{tab:runtime}, averaged over 100 runs of 17-frame generation. The main bottlenecks are the video model, which is amenable to distillation, and microfacet rendering, which can be sped up via lower-resolution visibility computation.
\begin{table}[h]
    \footnotesize 
    \centering
    \caption{
    \textbf{Runtime (ms).}
    \vspace{-6pt}
    }
    \label{tab:runtime}
    \begin{tabular}{|l|c|}
         \hline
         Stage & ms. \\
         \hline
         Microfacet & 404.6  \\
         RelightNet & 25.4  \\
         VAE env. & 140.2  \\
         Video model & 840.3  \\
         VAE dec. & 145.6  \\
         \hline
         Total & 1556.2 \\
         \hline
    \end{tabular}
    \vspace{-12pt}
\end{table}

\section{Ablations} \label{sec:s-ablations}

\begin{figure*}[h!]
\centering
    \includegraphics[page=1,trim={13.5cm 3.5cm 10.5cm 1cm},clip,width=0.9\textwidth]{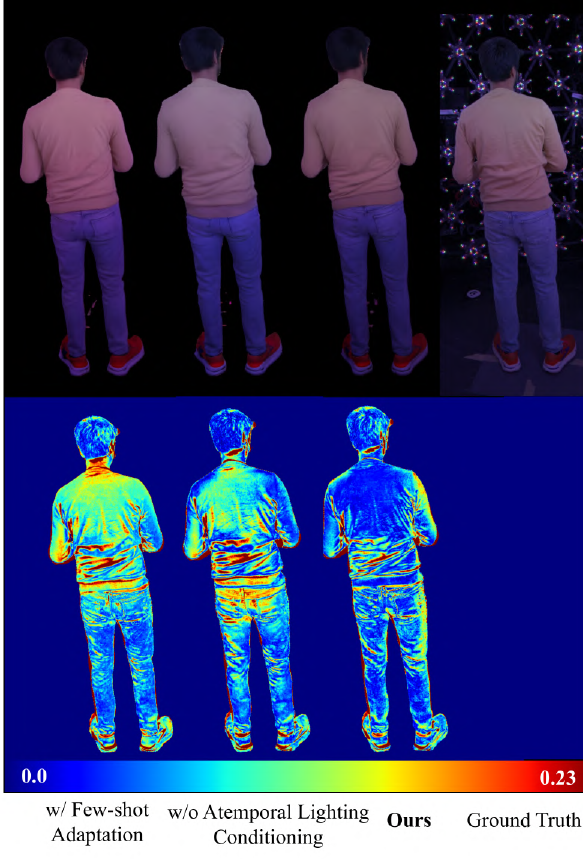}
    \caption{\textbf{Ablation of key design components.} The few-shot adaptation variant, while perceptually realistic, exhibits imprecise shading, while \emph{w/o Atemporal Lighting Conditioning} lacks global illumination context, resulting in inaccurate shading.}
    \label{fig:supp_comp}
\end{figure*}

\begin{figure*}[h!]
\centering
    \includegraphics[page=1,trim={9.5cm 6.45cm 1.5cm 1.2cm},clip,width=0.9\textwidth]{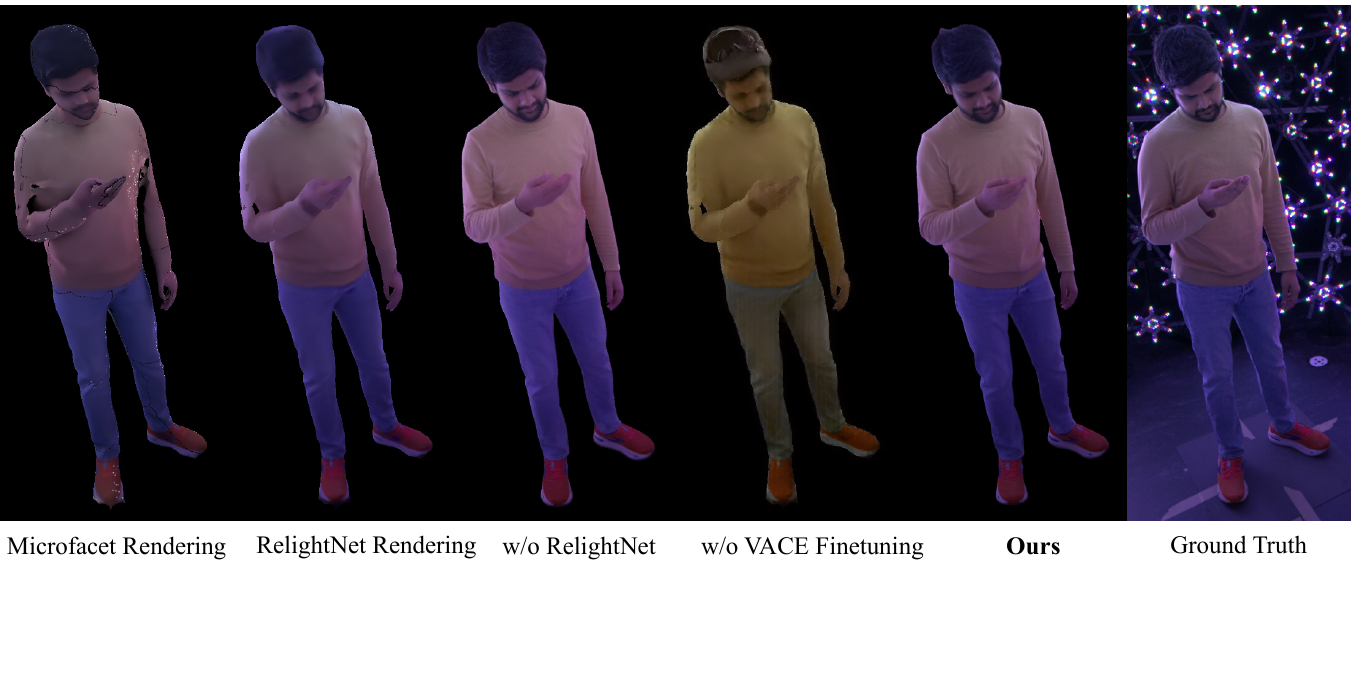}
    \vspace{-21pt}
    \caption{\textbf{Ablation of key design components.} Microfacet rendering exhibits artifacts due to simplified light transport assumptions. RelightNet improves global illumination but smooths high-frequency details due to pose-to-appearance ambiguity and mesh-based representation. Without RelightNet initialization, generative refinement produces incorrect shading. Without VACE finetuning, appearance degrades and lighting becomes physically implausible. } 
    \label{fig:exp-ablation}
\end{figure*}

We conduct quantitative (Tab.~2 of main paper) and qualitative (Fig.~\ref{fig:supp_comp}, Fig.~\ref{fig:exp-ablation}) ablation studies to validate each component of our method.

\par \noindent  \textbf{Physics-based Texture (Microfacet Rendering).}
The simplified BRDF, single-bounce assumption, and residual tracking inaccuracies prevent this stage from capturing pose-dependent appearance or complex global illumination, yielding coarse textures with visible approximation artifacts. Notably, it achieves reasonable PSNR (29.34) due to its physically correct low-frequency shading, confirming it provides a meaningful inductive bias for subsequent stages.

\par \noindent  \textbf{RelightNet Rendering.}
RelightNet improves upon the physics-based baseline by learning pose-dependent texture variations and expressive illumination effects. Interestingly, the regression-to-the-mean leads to better PSNR than our final model as blurrier predictions incur lower pixel-wise error, while SSIM improves because RelightNet learns coarse pose-correlated wrinkle patterns that bring predictions structurally closer to the ground truth. 
However, the deterministic nature of this process limits fine-scale appearance fidelity, as reflected in weaker LPIPS and FVD scores (Fig.~\ref{fig:exp-ablation}) compared to our final model with Generative Refinement. This trade-off motivates the generative refinement stage.

\par \noindent  \textbf{Ours w/ Temporal Lighting Conditioning.}
Replacing our atemporal cross-attention with full temporal cross-attention degrades performance (Tab.~2 of main paper). Since lighting changes every frame in our training data, temporal cross-attention must implicitly learn a near-diagonal attention matrix to associate each frame's features with its own environment map, while also attending over an attention matrix $T'$ times larger than its atemporal counterpart. Atemporal cross-attention induces this per-frame correspondence as an inductive bias, making the conditioning more sample-efficient. Using full temporal cross-attention also makes the network computationally heavier as it requires more GPU VRAM, necessitating workarounds like reduced envmap resolution or lesser cross-attention layers.

\par \noindent  \textbf{Ours w/o Atemporal Lighting Conditioning.}
This setting involves removing the illumination block from the generative refinement altoghther, leaving the WAN2.1-VACE model without any environment map conditioning.
Doing so leads to an expected drop in performance, demonstrating its necessity for reproducing fine-grained, temporally varying illumination effects. 
In the absence of illumination conditioning, the model can only extract the implicit lighting-specific signals from the coarse Relightnet output, which can only provide weak cues for relighting.
Our ablation analysis, especially the qualitative results (Fig.~\ref{fig:supp_comp}, Suppl. Vid.) demonstrate a noticeable shift in the character's appearance, due to the absence of an explicit illumination signal.

\par \noindent  \textbf{Ours w/o VACE.}
Replacing the VACE VCU with direct concatenation of conditioning inputs to noise latents leads to higher LPIPS and FVD, indicating that dense propagation of the conditioning signal throughout all model blocks is essential for effective spatio-temporal control. 
This is also consistent with the findings presented in the original VACE paper, which shows that VACE's controlnet-based conditioning is better than directly concatenating the reference to the denoising input.

\par \noindent  \textbf{Ours w/o RelightNet.}
Conditioning the generative stage directly on microfacet renderings leads to some shading inaccuracy, as the generative model must infer illumination effects without a physically informed intermediate representation. However, this variant retains direct access to the editable physics-based representation, enabling applications such as near-field relighting and texture editing where explicit material control is more important than shading refinement, as shown in the supplementary video.

\par \noindent  \textbf{Ours w/o VACE Finetuning.}
Without task-specific finetuning, the pretrained WAN2.1-VACE backbone hallucinates identity details, produces physically inconsistent shading, and drifts toward the reference appearance. We attribute this to the pretraining objective of VACE, which biases the model toward reproducing the appearance of the reference frame rather than transferring the lighting from a separate conditioning signal. Finetuning overcomes this bias by teaching the model to treat the coarse renderings as a shading target.

\par \noindent  \textbf{Ours w/o Prior.} 
Training the video model and the VACE layers from scratch fails to converge within the same budget (FVD: 394.91), confirming that the capture data alone is insufficient for the model to learn the full distribution of subject appearance in a reasonable training time. The pretrained video prior is essential to initialize the model in a meaningful region of this space.

\par \noindent  \textbf{Ours w/ Neural Gaffer (NG).}
Replacing our video model with image-to-image relighting model, Neural Gaffer~\cite{jin2024neural}, yields comparable per-frame relighting quality but severe temporal inconsistency, with high-frequency details varying arbitrarily across frames (which leads to higher FVD). For this, we finetune the complete Neural Gaffer model to accept coarse renderings $\boldsymbol{I}_t$ from RelightNet as input. This confirms that per-frame relighting quality alone is insufficient for coherent avatar relighting. The gains of our approach are primarily in the video rather than the image domain.

\par \noindent  \textbf{Ours w/ Few-shot Adaptation.}
Starting from a Subject~4 checkpoint, finetuning our video model on only ten 17-frame clips of Subject~1 for 200 iterations successfully generalizes to the new identity, demonstrating that subject-specific finetuning anchors appearance without catastrophically forgetting the broader appearance priors.

\par \noindent \textbf{Ours w/o Error Recycling.}
\begin{figure*}[h!]
\centering
    \includegraphics[trim={5cm 0 0 0},clip,width=1.0\linewidth]{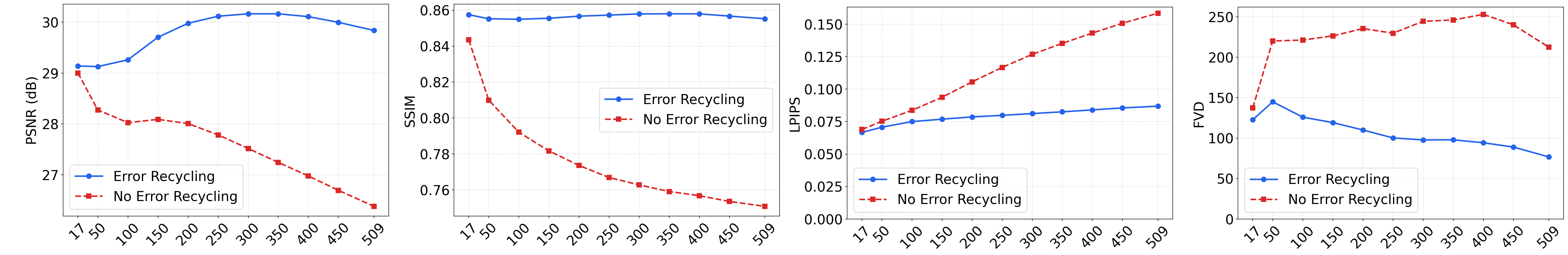}
    \caption{Metrics over clip length, w/ and w/o Error Recycling.}
    \label{fig:error_recycling}
\end{figure*}

We add a quantitative ablation on 509-frame
clips, generated every 1000 testing frames (Fig.~\ref{fig:error_recycling}). We report metrics on the first $k$ frames of each clip. Error Recycling results remains stable across clip lengths, while removing it causes consistent degradation, which is expected.
\begin{table}[t]
    \footnotesize 
    \centering
    \caption{
    \textbf{Additional Ablation Study.}
    We ablate the effect of tracking errors on our method. 
    \vspace{-6pt}
    }
    \label{tab:tracking}
    \begin{tabular}{|l|c|c|c|c|}
         \hline
         Method & PSNR $\uparrow$ & LPIPS  $\downarrow$ &SSIM $\uparrow$ &FVD $\downarrow$\\
         \hline
         2-view tracking & 26.25  & 11.23 & 81.85 & 73.49 \\
         4-view tracking & 29.53  & 7.12 & 86.28 & 52.69 \\
         \textbf{Ours} & \textbf{30.49}  & \textbf{6.42} & \textbf{87.52} & \textbf{48.79} \\
         \hline
    \end{tabular}
    \vspace{-12pt}
\end{table}

\par
\begin{figure*}[h!]
\centering
    \includegraphics[width=1.0\linewidth]{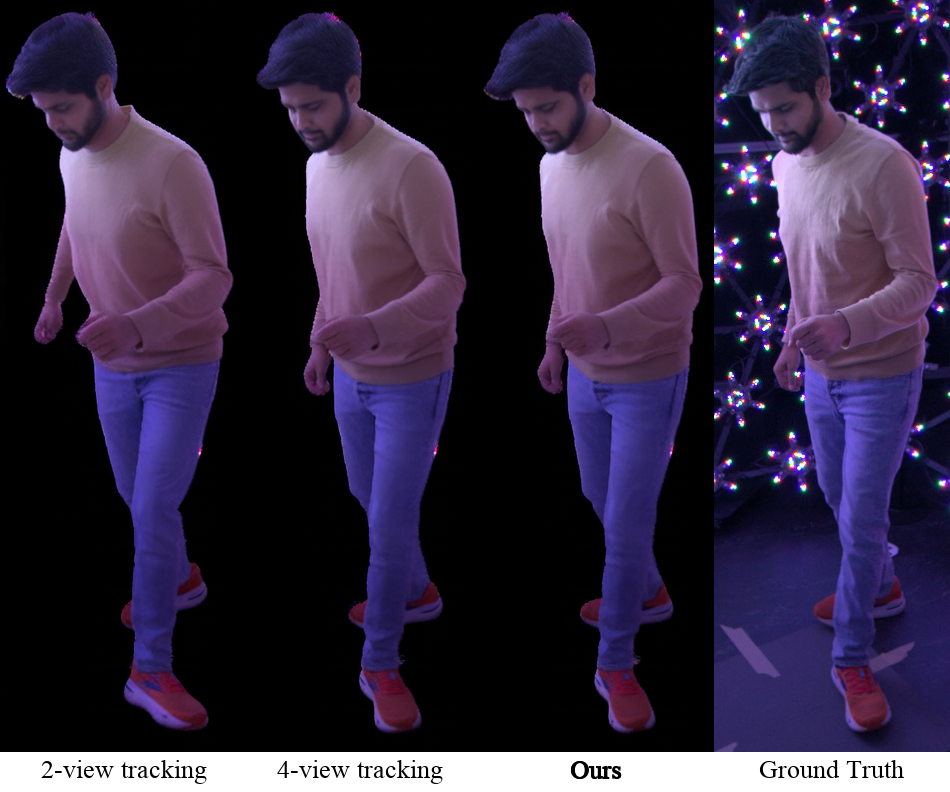}
    \caption{Qualitative ablation of tracking robustness.}
    \label{fig:tracking}
\end{figure*}

 \noindent \textbf{Ours w/ Tracking Error.}
We recompute test-time tracking from 4 and 2 views and compare it to dense-view tracking. Even as tracking worsens from 13.78 MPJPE (4-view) to 100.67 MPJPE (2-view), metrics degrade, but renderings remain coherent in most cases (Tab.~\ref{tab:tracking}, Fig.~\ref{fig:tracking}), suggesting robustness to noisier tracking.

\section{Details of Near-field Relighting} \label{sec:s-nfield}

For the near-field relighting experiment, we use our \emph{w/o RelightNet} model. In this setting, we generate physically based relighting using the same single-bounce approximation, but with a near-field point light source instead of an environment map, using our microfacet model. The resulting renderings are provided directly to the \emph{w/o RelightNet} model as conditioning input.

For the corresponding environment map conditioning, we assume a finite radius from the point light source and project rays from the center of the light stage toward the individual lights. If a ray intersects a light and subsequently does not intersect the subject, we assign an intensity to the corresponding environment map direction, corrected by the squared distance between the point light and the light stage emitter.

\section{Supplementary Video} \label{sec:s-video}
This section provides a detailed description of the supplementary video, including the content, and key results.

\subsection{Video Contents}
\begin{itemize}
    \item \textbf{0:00--0:48} --- Overview, with novel-view synthesis applied throughout the video. Demonstrations include dynamic relighting (0:17--0:24), camera manipulation (0:25--0:34), wrinkle diversity (0:37--0:44), and long-form video generation.
    
    \item \textbf{0:49--1:07} --- Generalization to out-of-distribution OLAT lighting conditions, highlighting self-shadowing effects.
    
    \item \textbf{1:09--1:21} --- Qualitative evaluation of a model variant enabling near-field relighting without retraining, contrasted with far-field relighting. Occlusion effects of the light source by the subject's arm are also illustrated.
    
    \item \textbf{1:22--1:28} --- Relightable texture editing results produced using the aforementioned variant.
    
    \item \textbf{1:30--1:59} --- Detailed visualization of the method pipeline, including skeleton representation, DDC mesh, microfacet-based rendering, RelightNet output, and final denoising to produce the refined frames.
    
    \item \textbf{2:00--2:16} --- Qualitative results for Subject 4, showcasing spiral novel-view synthesis across multiple stages of the model.
    
    \item \textbf{2:17--2:25} --- Qualitative results for Subject 3, illustrating arcing novel-view synthesis under an outdoor environment map.
    
    \item \textbf{2:26--2:37} --- Qualitative comparisons against R4D~\cite{chen2022relighting4d}, IntrinsicAvatar~\cite{wang2024intrinsicavatar}, and MeshAvatar~\cite{chen2024meshavatar}. These deterministic methods, employing either implicit radiance fields or hybrid implicit/explicit representations, fail to capture high-frequency details that our approach preserves.
    
    \item \textbf{2:39--2:48} --- Qualitative comparisons against 3D anisotropic Gaussian based DNF-Avatar~\cite{jiang2025dnf} and RHC~\cite{anon2025rhc} (0-view and 4-view variants). Our method outperforms all baselines, including sparse-view RHC, in terms of visual fidelity and detail preservation.
    
    \item \textbf{2:49--3:04} --- Ablation studies highlighting the effect of key components: \emph{w/o Error Recycling} introduces artifacts in long-form sequences, \emph{w/o RelightNet} produces incorrect shading, and \emph{w/o VACE Finetuning} results in deviation from the coarse relit input, especially in extended sequences.
    
    \item \textbf{3:05--3:10} --- Additional ablations: \emph{w/ Neural Gaffer} achieves high frame quality but exhibits temporal instability; \emph{w/ Few-shot Adaptation} yields perceptually plausible relighting with some shading inaccuracies; \emph{w/o Atemporal Lighting Conditioning} lacks global illumination context, producing incorrect shading.
\end{itemize}

\subsection{Visualization Details}
The supplementary video contains several visualizations rendered with a background environment, and a virtual floor with rendered shadows. 
This contextualizes our work with respect to the overall task of character relighting.
As performing such visualization involves post-processing details, we provide them in the following paragraphs.
\par \noindent \textbf{Background Generation.}
To synthesize a background consistent with the illumination used for relighting, we project the environment map into the camera view. Specifically, the environment map is first transformed from its native parameterization into world space and subsequently projected into camera space using the \emph{skylibs} library~\cite{skylibs}. This produces a background image that is spatially aligned with the rendered subject. The focal length $f$ used for the projection is selected empirically to achieve visually consistent perspective and perceptual integration of the subject with the background across different scenes.

\par \noindent \textbf{Platform and Shadow Generation.}
To provide additional spatial context, we place the subject on a simple geometric platform created in \emph{Blender}. The environment map is converted to Blender’s coordinate system and used as the illumination source for rendering. We then place the animated mesh $\boldsymbol{M}_t$ on the platform and render the scene under this lighting. This procedure produces shadows cast by the subject onto the platform that are consistent with the environment illumination, improving the realism of the final composite.

\par \noindent \textbf{Foreground Segmentation.}
Our method generates relit frames with the subject rendered on a black background. To enable compositing with the generated background and shadow layers, we extract a foreground mask for the subject. We perform this segmentation using MatAnyone2~\cite{yang2026matanyone2}, which provides temporally stable masks across frames. The resulting masks allow clean compositing of the foreground subject with the background and shadow renderings.
\begin{table}[t]
    \footnotesize 
    \centering
    \caption{
    \textbf{Quantitative Comparison with RHC.}
    We compare our method with the sparse-view Gaussian avatar method RHC and its pose-only variant on Subject~1. Our method achieves lower LPIPS and FVD while requiring no image conditioning at test time.
    }
    \label{tab:rhc}
    \begin{tabular}{|l|c|c|c|c|}
         \hline
         Method & PSNR $\uparrow$ & LPIPS $\downarrow$ & SSIM $\uparrow$ & FVD $\downarrow$\\
         \hline
         RHC~\cite{anon2025rhc} (4 views) & \textbf{31.38} & 7.01 & \textbf{90.00} & 54.57 \\
         RHC~\cite{anon2025rhc} (0 views) & 30.75 & 7.84 & 88.32 & 85.75 \\
         \hline
         \textbf{Ours (0 views)} & 30.49 & \textbf{6.42} & 87.52 & \textbf{48.79} \\
         \hline
    \end{tabular}
\end{table}
\par \noindent \textbf{Flicker Correction.}
While error-recycling produces long-form videos with high-frequency detail, temporal flicker can occasionally occur when the network misinterprets shading from previous frames as error.
We correct this using an edge-aware guided image filter~\cite{he2010guided} with a temporally coherent coarse rendering $\boldsymbol{I}_t$ from RelightNet as a guide. We first apply the guided filter to both the coarse rendering and the input frame using the input as the guide in both cases. This extracts their respective low-frequency illumination maps that are aligned to the edge structure of the input. We then compute the residual between these two illumination maps and additively blend it back into the input frame, effectively nudging its illumination toward the stable temporal profile of the coarse rendering while preserving its fine-grained detail.

We note that none of the post-processing steps discussed above are used to compute the results or metrics reported in the paper.
\vspace{-11pt}

\section{Comparison with Recent Gaussian-Based Relighting Methods} \label{sec:s-rhc}

\begin{figure*}[t]
    \centering \includegraphics[trim={8cm 3cm 11cm 0cm}, clip,width=0.9\linewidth]{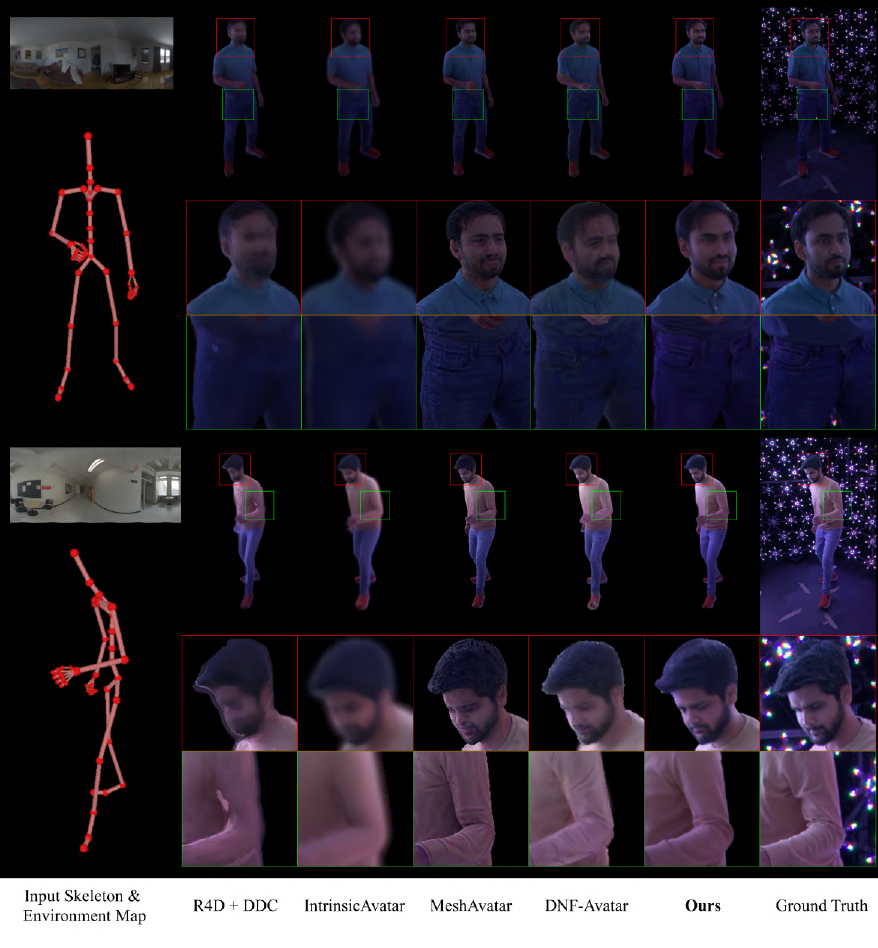}
    \caption{Qualitative comparison against baseline methods in the testing sequence of additional two subjects.}
    \label{fig:s-more-results}
\end{figure*}
We compare our approach to Relightable Holoported Characters (RHC)~\cite{anon2025rhc}, a \textbf{sparse-view} driven method for novel-view synthesis and relighting that requires four input views of the subject at test time. We additionally evaluate a pose-only variant of RHC that does not use image conditioning. To obtain this variant, we remove the Albedo and Sapiens features from its physics-informed feature stack and retrain the model.

In contrast to the sparse-view driven setting of RHC, our method does not require any image inputs at test time and is purely pose-driven. Despite this, Table~\ref{tab:rhc} shows that our approach outperforms both the 4-view and 0-view variants of RHC in terms of LPIPS and FVD. We show qualitative results in the supplementary video. This suggests that the prior learned by the video diffusion model aligns more closely with real-world appearance statistics than the subject-specific model used in RHC, which is trained only on the data of a single subject.

\vspace{-8pt}
\section{Additional Qualitative Results}\label{sec:s-qual}
We show additional qualitative results in Fig.~\ref{fig:s-more-results}, where our method constantly achieves superior results with sharper clothing and facial details, with more accurate color shading.

\section{Limitations and Future Work} \label{sec:s-limits}
While GRA enables photorealistic 3D full-body relightable avatars, it has a few limitations that point to promising directions for future work.
First, our use of error recycling for long-horizon generation can introduce minor temporal color jitter, as the model may imperfectly disentangle shading variations from injected error during continuation.
This artifact could be mitigated by incorporating a lightweight post-processing deflickering module, such as blind video deflickering networks~\cite{lei2023blind}.
Second, our current implementation supports rendering at 2 seconds/frame.
Future work could substantially improve runtime performance by distilling the generative and relighting components into more compact models~\cite{hinton2015distilling}, enabling real-time or interactive frame rates.
Finally, our method focuses primarily on full-body appearance and does not explicitly model fine-grained facial expressions or hand articulation.
Integrating dedicated high-fidelity face~\cite{sanyal2019learning} and hand~\cite{romero2022embodied} models would further enhance expressiveness and visual realism in these regions.

\end{document}